\documentclass[twoside]{article}

\usepackage[preprint]{aistats2025}
\usepackage[round]{natbib}

\usepackage{url}
\usepackage{xcolor}
\definecolor{aistatsblue}{RGB}{0, 20, 115}

\usepackage[colorlinks=true,
            citecolor=aistatsblue,
            linkcolor=aistatsblue,
            urlcolor=aistatsblue]{hyperref}
            
\usepackage{algorithm}
\usepackage{algorithmic}
\usepackage{amsfonts}
\usepackage{amsmath}
\usepackage{amsthm}
\usepackage{amssymb}

\newcommand{\mc}{\mathcal}
\newcommand{\mb}{\mathbb}

\usepackage{amsmath}

\usepackage{booktabs}
\usepackage{graphicx}
\usepackage{thmtools, thm-restate} % For restatable theorems
\declaretheorem[name=Theorem,numberwithin=section]{theorem}
\declaretheorem[name=Proposition,sibling=theorem]{proposition}

\declaretheorem[name=Definition,sibling=theorem]{definition}

\declaretheorem[style=remark,name=Remark]{remark}
\begin{document}


% \aistatsauthor{ Yannis Montreuil* \And Letian Yu* \And  Axel Carlier \\
% \And Lai Xing Ng \And Ooi Wei Tsang }

\runningtitle{Adversarial Robustness in One-Stage Learning-to-Defer}
\runningauthor{Montreuil et al.}

\twocolumn[

\aistatstitle{Adversarial Robustness in One-Stage Learning-to-Defer}

\aistatsauthor{
Yannis Montreuil$^{*,1,4,5}$ \And
Letian Yu$^{*,1}$ \And
Axel Carlier$^{2,4}$ \AND
Lai Xing Ng$^{3,4}$ \And
Wei Tsang Ooi$^{1,4}$
}

\aistatsaddress{
$^{1}$School of Computing, National University of Singapore, Singapore \\
$^{2}$Fédération ENAC ISAE-SUPAERO ONERA, Université de Toulouse, France \\
$^{3}$Institute for Infocomm Research, A*STAR , Singapore \\
$^{4}$IPAL, IRL 2955, Singapore \\
$^{5}$CNRS@CREATE LTD, 1 Create Way, Singapore
}

]

\begin{abstract}
Learning-to-Defer (L2D) enables hybrid decision-making by routing inputs either to a predictor or to external experts. While promising, L2D is highly vulnerable to adversarial perturbations, which can not only flip predictions but also manipulate deferral decisions. Prior robustness analyses focus solely on two-stage settings, leaving open the end-to-end (one-stage) case where predictor and allocation are trained jointly. We introduce the first framework for adversarial robustness in one-stage L2D, covering both classification and regression. Our approach formalizes attacks, proposes cost-sensitive adversarial surrogate losses, and establishes theoretical guarantees including $\mathcal{H}$, $(\mathcal{R }, \mathcal{F})$, and Bayes consistency. Experiments on benchmark datasets confirm that our methods improve robustness against untargeted and targeted attacks while preserving clean performance. 
\end{abstract}

\section{INTRODUCTION}
With the growing adoption of routing-based methods, 
\emph{Learning-to-Defer} (L2D) has emerged as a principled framework for hybrid decision-making~\citep{madras2018predict, mozannar2021consistent}. 
An L2D system may either predict directly or defer to an expert, thereby trading off predictive performance against selective reliance on external decision-makers. 
This framework is particularly relevant in safety-critical applications~\citep{joshi2021learning, strong2025trustworthy, palomba2025a} 
and also offers a unifying perspective on a broad class of routing problems~\citep{chenfrugalgpt, jitkrittum2025universal}, 
where the central challenge is to design allocation policies that distribute decisions optimally across multiple agents.

Yet, despite its promise, L2D inherits the adversarial vulnerabilities of standard 
machine learning models \citep{goodfellow2014explaining, Madry2017TowardsDL, jia2017adversarial} while introducing new ones unique to the routing setting. 
Recent work~\citep{montreuil2025adversarial} 
shows that L2D systems can be even more fragile to adversarial perturbations 
than conventional systems. An adversary may not only induce misclassification but also manipulate whether, and to whom, 
the system defers—for example, bypassing a reliable expert or forcing deferral to one 
known to perform poorly. 
Such attacks threaten both predictive performance and the reliability of decision allocation, 
raising pressing concerns for deployment in safety-critical environments.

Existing robustness studies focus exclusively on the simplified \emph{two-stage} L2D setting, 
where the predictor and experts are trained offline and only the allocation policy is learned~\citep{Narasimhan, mao2023twostage, mao2024regressionmultiexpertdeferral, montreuil2024twostagelearningtodefermultitasklearning, montreuil2025adversarial, montreuil2025askaskktwostage}. 
This formulation is convenient but fails to capture the complexity of end-to-end (\emph{one-stage}) L2D, 
in which the predictor and allocation policy must be optimized jointly. 
We address two problems: (i) characterizing attacks that target one-stage L2D systems, and (ii) designing one-stage L2D methods with guarantees for the proposed adversarial losses.

In this work, we develop a framework for adversarial robustness in one-stage L2D, encompassing both classification and regression. 
Our contributions are threefold: 

\begin{enumerate}
    \item We introduce both untargeted and targeted attacks that reveal how adversaries can manipulate prediction and deferral simultaneously;
    \item We propose new outcome-wise adversarial surrogate deferral losses with tractable relaxations and establish theoretical guarantees for the proposed adversarial losses, including $\mathcal{H}$-consistency for classification and $(\mathcal{R},\mathcal{F})$-consistency for regression;
    \item We demonstrate empirically, on image classification and tabular regression benchmarks, that our methods substantially improve robustness against both attack types while maintaining competitive clean accuracy. 
\end{enumerate}

Together, these results provide a theoretically grounded and practically effective approach to adversarially robust one-stage L2D.

\section{RELATED WORK} \label{related}

\textbf{One-Stage L2D.}
\citet{madras2018predict} introduced the first formal framework for Learning-to-Defer (L2D), incorporating expert predictions via a defer-to-expert mechanism. A key advance was made by \citet{mozannar2021consistent}, who proposed a \emph{score-based} formulation of L2D. In this approach, the classifier is augmented with a shared scoring function that jointly determines both prediction and allocation, thereby unifying two decisions that had previously been treated separately.

Subsequent work has pursued several directions: strengthening theoretical guarantees such as $\mathcal{H}$-consistency and realizability \citep{Mozannar2023WhoSP, mao2024principledapproacheslearningdefer, mao2024realizablehconsistentbayesconsistentloss, mao2025mastering, mao2025thesis}; broadening the class of statistically sound surrogate losses \citep{charusaie2022sample, mao2024principledapproacheslearningdefer, montreuil2026beyond, montreuil2026learningtodeferexpertconditionedadvice}; improving estimation methods \citep{Verma2022LearningTD, verma_multiple, Cao_Mozannar_Feng_Wei_An_2023}; addressing budgeted, imbalanced, or non-stationary deferral regimes \citep{desalvo2025budgeted, cortes2026optimized, montreuil2026learningdefernonstationarytime}; and extending policies to top-$k$ expert selection \citep{montreuil2025askaskktwostage}. The score-based methodology has also been applied across diverse classification domains \citep{Keswani, Kerrigan, Hemmer, Benz, Tailor, liu2024mitigating, montreuil2025optimalqueryallocationextractive, montreuil2026online}. Regression-based variants have been proposed by \citet{mao2024regressionmultiexpertdeferral}, who employ a dedicated allocation policy alongside a trainable predictor.
\textbf{Robustness in L2D.}
Robustness has received comparatively little attention in the L2D literature. To the best of our knowledge, the only prior work addressing robustness is that of \citet{montreuil2025adversarial}, who introduced an adversarially consistent formulation. Their analysis, however, is restricted to the two-stage setting, where experts are fixed and not jointly optimized with the allocation policy. This assumption simplifies the problem: robustness reduces to modifying the allocation strategy without accounting for the interaction between learning the predictor and adapting to the experts.

By contrast, the one-stage setting presents a qualitatively harder challenge as it requires jointly learning the predictor as well as the allocation policy. In this work, we extend the robustness framework of \citet{montreuil2025adversarial} to the technically more demanding one-stage L2D scenario, where adversarial perturbations may influence all components of the system simultaneously.
\section{PRELIMINARIES}

\paragraph{Task and data.}
Let $\mathcal{X}\subseteq\mathbb{R}^d$ be the input space and let $\mathcal{Z}$ denote the target space, with
\[
\mathcal{Z} \;=\;
\begin{cases}
\mathcal{Y}=\{1,\dots,K\}, & \text{classification},\\
\mathcal{T}\subseteq\mathbb{R}^m, & \text{regression}.
\end{cases}
\]
Each training label satisfies $z \in \mathcal{Z}$.  
We assume access to i.i.d.\ samples $\mathcal{S}_n = \{(x_i, z_i, \mathbf{m}_i)\}_{i=1}^n
    \sim \big(\mathcal{D} \times \mathcal{M}\big)^n$, where $(x,z) \sim \mathcal{D}$ is drawn from the underlying data distribution and 
$\mathbf{m} \sim \mathcal{M}\!\mid(x,z)$ denotes the outputs of the $J$ fixed experts 
conditioned on $(x,z)$ \citep{madras2018predict, mozannar2021consistent, Verma2022LearningTD}. Here $\mathbf{m} = (m_{1}, \dots, m_{J})$, with each 
$m_{j}$ belonging to the appropriate output space: in classification, 
$m_{j} \in \mathcal{Y}$ is a categorical prediction, while in regression, 
$m_{j} \in \mathcal{T}$ is a real-valued prediction. All expert predictions can thus be regarded as elements of the common target space $\mathcal{Z}$.

\textbf{Classification L2D.}
We specialize to $\mathcal{Z}=\mathcal{Y}=\{1,\dots,K\}$. Each training label $y_i\in\mathcal{Y}$ and each expert output $m_{ij}\in\mathcal{Y}$ is categorical. Alongside the $J$ experts $m_1,\dots,m_J$, we train a hypothesis $h\in\mathcal{H}$ with $h:\mathcal{X}\to\mathbb{R}^{K+J}$, where the first $K$ coordinates correspond to class scores and the last $J$ to expert scores \citep{mozannar2021consistent, Cao_Mozannar_Feng_Wei_An_2023, mao2024principledapproacheslearningdefer, montreuil2025askaskktwostage}. Let $\mathcal{A}^{c} = \mathcal{Y} \cup \{K+1,\dots,K+J\}$ denote the augmented action space. 
For a vector $\xi(x)$, we write $\hat{\xi}(x)$ to denote its $\arg\max$ index. 
The induced decision rule is then $\hat{h}(x) = \arg\max_{k \in \mathcal{A}^{c}} h(x)_k $, with ties broken uniformly at random.
 If $\hat{h}(x) \in \mathcal{Y}$, the system predicts the corresponding category; 
if $\hat{h}(x) = K+j$ for some $j \in \{1,\dots,J\}$, the system defers to expert $m_j$. 

The predictor $h \in \mathcal{H}$ is trained to minimize the expected risk of the \emph{true deferral loss} for classification
\begin{equation}\label{class}
\begin{aligned}
    \mspace{-8mu}\ell_{\text{def}}^c(\hat{h}(x),y, \mathbf{m}) \mspace{-3mu} = \mspace{-3mu}
\begin{cases}
\mathbf{1}\{\hat{h}(x)\neq y\}, \mspace{-10mu} & \text{if } \hat{h}(x)\in\mathcal{Y},\\[0.3em]
c^c_j(m_j,y), & \text{if } \hat{h}(x)=K+j.
\end{cases}
\end{aligned}
\end{equation}
where $c^c_j(m_j,y) \in [0,1]$ is the cost of deferring to expert $m_j$. We set $c^c_j(m_j,y) \;=\; \alpha_j\,\mathbf{1}\{m_j\neq y\} + \beta_j$, 
with $\alpha_j\ge 0$ a scaling coefficient and $\beta_j$ the fixed consultation cost for querying expert $j$. 

A common way to minimize such a discontinuous loss is to replace it with a consistent surrogate
\citep{Steinwart2007HowTC, Statistical, bartlett1, Awasthi_Mao_Mohri_Zhong_2022_multi, mao2024hconsistencyregression, mao2024multilabel, mao2024universalgrowth, mao2025enhanced, zhong2025thesis, mohri2026beyond, cortes2025improvedbalanced, cortes2025balancingscales, mao2025principledbinary, mohri2026linear, mohri2026mind},
ensuring that the learned predictor approximates the Bayes classifier $h^B \in \mathcal{H}$ that minimizes the expected true deferral loss. 
Given the augmented action space $\mathcal{A}^c$, we define the \emph{surrogate deferral loss} for classification as
\begin{equation}
\label{eq:surr_classification}
\begin{aligned}
\Phi_{\text{def}}^{c, u}(h&(x),y, \mathbf{m}) \; = \; \Phi_{\text{cls}}^u(h(x),y) \\
& +\sum_{j=1}^J (1-c_j^c(m_j,y))\,\Phi_{\text{cls}}^u(h(x),K+j),
\end{aligned}
\end{equation}
where $\Phi_{\text{cls}}^u$ is a classification surrogate from the cross-entropy family 
\citep{mao2023crossentropylossfunctionstheoretical}, defined as $\Phi_{\text{cls}}^u(h(x),y)
=\Psi^u\!\Big(\sum_{y'\in\mc{Y}}\exp(h(x)_{y'}-h(x)_y) -1\Big)$
with outer transform
\begin{equation}
    \Psi^u(v)=
\begin{cases}
\log(1+v), & u=1,\\[0.3em]
\tfrac{1}{1-u}\big[(1+v)^{1-u}-1\big], & u\neq 1.
\end{cases}
\end{equation}
This family recovers well-known surrogates, including logistic loss \citep{Ohn_Aldrich1997-wn}, generalized cross-entropy \citep{zhang2018generalizedcrossentropyloss}, and mean absolute error \citep{Ghosh}. The surrogate risk is $\mc{E}_{\Phi_{\text{def}}^{c, u}}(h)=\mb{E}[\Phi_{\text{def}}^{c,u}(h(X), Y, M)]$ with optimal value $\mc{E}_{\Phi_{\text{def}}^{c, u}}^\ast(\mc{H})=\inf_{h\in\mc{H}}\mc{E}_{\Phi_{\text{def}}^{c, u}}(h)$. A surrogate is called \emph{consistent} if minimizing its excess risk also minimizes the true excess risk \citep{Statistical, bartlett1, Steinwart2007HowTC, tewari07a}.

\begin{theorem}[$\mc{H}$-consistency bounds \citep{Awasthi_Mao_Mohri_Zhong_2022_multi}]
The surrogate $\Phi_{\text{def}}^{c, u}$ is $\mc{H}$-consistent with respect to $\ell_{\text{def}}^{c}$ if there exists a non-decreasing function $\Gamma^u:\mb{R}^+\to\mb{R}^+$ such that for every distribution $\mc{D}$,
\begin{equation*}
    \begin{aligned}
        &\mc{E}_{\ell_{\mathrm{def}}^c}(h)-\mc{E}_{\ell_{\mathrm{def}}^c}^B(\mc{H})+\mc{U}_{\ell_{\mathrm{def}}^c}(\mc{H})  \leq \\
    & \qquad \Gamma^u\!\Big(\mc{E}_{\Phi_{\mathrm{def}}^{c, u}}(h)-\mc{E}_{\Phi_{\mathrm{def}}^{c, u}}^\ast(\mc{H})+\mc{U}_{\Phi_{\mathrm{def}}^{c, u}}(\mc{H})\Big).
    \end{aligned}
\end{equation*}
\end{theorem}

Here $\mc{U}_{\Phi_{\mathrm{def}}^{c, u}}(\mc{H})$ is the \emph{minimizability gap}, quantifying the discrepancy between the best achievable excess risk within $\mc{H}$ and the expected pointwise minimum. This gap vanishes when $\mc{H}=\mc{H}_{\mathrm{all}}$, recovering Bayes-consistency in the asymptotic limit \citep{Statistical, Steinwart2007HowTC}. Beyond the classification setting studied here, this hypothesis-dependent notion of $\mc{H}$-consistency has been developed into a broad theory~\citep{Awasthi_Mao_Mohri_Zhong_2022_multi, mao2024h}, with guarantees established for multi-class abstention and deferral~\citep{theoretically, Mao_Mohri_Zhong_2023, mohri2024learningreject}, for ranking and cardinality-aware prediction~\citep{mao2023pairwisemisranking, mao2023rankingabstention, cortes2024cardinalityaware}, and for structured prediction~\citep{mao2023structuredprediction}, as well as in recent algorithms for generalized-metric optimization and robust generative modeling~\citep{mohri2026generalized, mohri2026principled, cortes2026theoretical}.

\textbf{Regression L2D.} 
Let $\mathcal{Z}=\mathcal{T}\subseteq\mathbb{R}^m$. 
Each training label $t_i\in\mathcal{T}$ and each expert output $m_{ij}\in\mathcal{T}$ is a real-valued vector.  The system consists of a predictor $f\in\mathcal{F}$, with $f:\mathcal{X}\to\mathcal{T}$, and a rejector $r\in\mathcal{R}$, with $r:\mathcal{X}\to\mathbb{R}^{J+1}$. 
Let $\mathcal{A}^{r}=\{1,2,\dots,J+1\}$ denote this action space. 
The induced decision rule is $\hat{r}(x)= \arg\max_{k\in\mathcal{A}^{r}} r(x)_k$, 
so that if $\hat{r}(x)=1$, the system outputs $f(x)$, while if $\hat{r}(x)=j$ with $j\ge 2$, the system defers to expert $m_{j-1}$. 
The \textit{true deferral loss} for regression is
\begin{equation}\label{regression}
\ell_{\text{def}}^r(f(x),\hat{r}(x),t, \mathbf{m}) 
\mspace{-3mu}=\mspace{-3mu} \sum_{j=1}^{J+1} \mspace{-3mu}c_j^r(f(x), m_{j-1}, t)\mspace{-1mu}\mathbf{1}\{\hat{r}(x)=j\},
\end{equation}
where each $c_j^r(f(x), m_{j-1}, t)\ge 0$ is the cost of action $j$. 
For the predictor we define $c_1^r(f(x),t)=\alpha_1\,L(f(x),t)+\beta_1$, 
and for expert $j\ge 2$, $c_j^r(m_{j-1},t)=\alpha_j\,L(m_{j-1},t)+\beta_j$, 
where $L:\mathcal{T}\times\mathcal{T}\to\mathbb{R}_+$ is a regression loss (e.g.\ squared error). Similarly to the classification case, we approximate the discontinuous loss with a consistent surrogate, which we refer to as the \emph{surrogate deferral loss} for regression:
\begin{equation}
\begin{aligned}
\Phi_{\text{def}}^{r,u}(f(x),r(x),t, \mathbf{m}) 
&= \sum_{j=1}^{J+1} \tau_j^r(f(x), \mathbf{m}, t)\,\Phi_{\text{cls}}^u(r(x),j) \\
&\qquad - (J-1)\,c_1^r(f(x),t),
\end{aligned}
\end{equation}
where $\Phi_{\text{cls}}^u$ is the same cross-entropy–type surrogate used in classification, 
and the weights $\tau_j^r(f(x),\mathbf{m}, t)$ are defined as $\tau_j^r(f(x),\mathbf{m}, t)=\sum_{i\not= j}c_i^r(f(x), m_{i-1}, t)$. 
This surrogate has been shown to be both Bayes consistent and $(\mathcal{R},\mathcal{F})$-consistent \citep{mao2024regressionmultiexpertdeferral}.

\paragraph{Robustness.} 
Adversarially robust classification aims to train classifiers that remain reliable under small, often imperceptible, perturbations of the input \citep{goodfellow2014explaining, Madry2017TowardsDL}. The goal is to minimize the \emph{true multiclass loss} $\ell_{01}$ evaluated on an adversarial input $x' = x + \delta$ \citep{Gowal2020UncoveringTL, Awasthi_Mao_Mohri_Zhong_2022_multi}. A perturbation $\delta$ is constrained by its magnitude, with the adversarial region around $x$ defined as $B_p(x,\gamma) = \{\, x' \in \mathbb{R}^d \mid \|x' - x\|_p \leq \gamma \,\}$, where $\|\cdot\|_p$ denotes the $p$-norm and $\gamma > 0$ bounds the perturbation size. 

The \emph{adversarial $0$–$1$ loss} is $\widetilde{\ell}_{01}(h,x,y) = \sup_{x' \in B_p(x,\gamma)} \mathbf{1}\{\hat{h}(x') \neq y\}$, and its \emph{adversarial margin surrogate} is 
\begin{equation}
    \widetilde{\ell}_{01}(h,x,y)\leq \sup_{x' \in B_p(x,\gamma)} \Phi_{\text{cls}}^{\rho, u}(h(x'),y),
\end{equation}
with $\Phi_{\text{cls}}^{\rho, u}(h(x'),y)=\Psi^u\!\Big(\sum_{k \neq y} \Psi_\rho\big(h(x')_k - h(x')_y\big)\Big)$. Here, $\Psi^u$ and $\Psi_\rho$ are transformations that characterize the surrogate family. In the analysis below, we use the exponential $\rho$-soft margin transform $\Psi_\rho(v)=\exp(-v/\rho)$. Recent works show that algorithms based on smooth, regularized variants of these comp-sum $\rho$-margin losses achieve strong calibration and consistency guarantees \citep{Awasthi_Mao_Mohri_Zhong_2022_multi, Grounded, mao2023crossentropylossfunctionstheoretical}.

\section{ATTACKING ONE-STAGE LEARNING-TO-DEFER APPROACHES}
As discussed in Section~\ref{related}, robustness has so far been investigated only in the two-stage setting of Learning-to-Defer \citep{montreuil2025adversarial}, while the one-stage formulation remains unexplored. Addressing robustness in this setting requires novel surrogate losses that simultaneously preserve consistency and provide robustness guarantees.

Learning-to-Defer seeks to route each query to the most reliable agent—either the predictor or one of the experts \citep{madras2018predict, mozannar2021consistent}. In classification, the augmented classifier jointly handles both prediction and deferral. In regression, allocation and prediction are decoupled. Such formulations perform well on clean inputs \citep{verma_multiple, Mozannar2023WhoSP}. The natural question is: do they remain effective under noisy or adversarial perturbations? We answer this in the negative. Extending the conclusions of \citet{montreuil2025adversarial}, we show that adversarial attacks not only compromise the allocation policy but also predictive performance in both the regression and classification settings of one-stage Learning-to-Defer. 

\subsection{Introducing Attacks}

\paragraph{Untargeted Attack.} 
In the \emph{untargeted} setting, the adversary seeks a perturbation 
$\delta \in B_p(0,\gamma)$ that maximally degrades the system’s performance, 
without targeting a specific outcome. 

\begin{restatable}[Untargeted Attack]{definition}{untargeted}\label{def:untargeted}
An \emph{untargeted adversarial attack} in L2D is the problem of finding a perturbed input 
$x' = x + \delta$, with $\delta \in B_p(0,\gamma)$, that maximizes a differentiable attack objective aligned with the deferral loss 
in either classification or regression. In our experiments, we instantiate this objective with the corresponding surrogate loss. Formally,
\[
    x' = \underset{x' \in B_p(x,\gamma)}{\arg\max}\;
    \Phi_{\mathrm{def}}^{\bullet,u}(x'),
\]
where $\bullet \in \{c,r\}$ indicates whether the task is classification ($c$) or regression ($r$).
\end{restatable}

The attacker perturbs an input $x$ so that, even if the clean input would be processed optimally, 
the perturbed version $x'$ induces a worse outcome under the deferral policy. 
Such degradation can occur in several ways in the one-stage setting:  
(i) causing the predictor to output an incorrect class in classification or a high-error estimate in regression;  
(ii) obstructing deferral to a reliable expert when such deferral would reduce error; or  
(iii) inducing unnecessary deferral when the predictor is already sufficiently accurate.  
For instance, if the clean input $x$ should be handled by the main predictor, 
an untargeted perturbation may instead redirect the query to an inappropriate expert, 
thereby increasing the overall system error.

\paragraph{Targeted Attack.} 
In the \emph{targeted} setting, the adversary specifies a desired outcome $\nu$ 
and perturbs the input so that the system’s decision is steered toward this target. 
In classification, $\nu$ may correspond to a class label, $\nu \in \{1,\dots,K\}$, 
or to a deferral action, $\nu \in \{K+1,\dots,K+J\}$. 
In regression, the target space is defined analogously: 
$\nu = 1$ denotes trusting the predictor’s output, while $\nu \in \{2,\dots,J+1\}$ 
corresponds to deferral to expert $\nu$. 
Let $\boldsymbol{\delta} = (\delta_1,\dots,\delta_{|\mc{A}|})$ denote the 
family of outcome-specific perturbations, with $\delta_j \in \mc{X}$.

\begin{restatable}[Targeted Attack]{definition}{targeted}\label{def:targeted}
A \emph{targeted adversarial attack} in L2D is the problem of finding a perturbation 
$\delta_\nu \in \mc{X}$ from the family $\boldsymbol{\delta}$ 
such that the adversarial input $x'_\nu = x + \delta_\nu$ lies within the $p$-norm ball 
$B_p(x,\gamma)$ and drives the allocation policy toward the specified target $\nu \in \mc{A}$. 
Formally,
\[
    x'_\nu = \underset{x'_\nu \in B_p(x,\gamma)}{\arg\min}\;
    \Phi_{\mathrm{cls}}^u\big(\pi(x_\nu'), \nu\big),
\]
where $\pi$ is $h$ in classification or $r$ in regression. 
\end{restatable}

Unlike untargeted attacks, which aim only to degrade performance, 
targeted attacks explicitly redirect the decision to a chosen outcome $\nu$. 
The surrogate $\Phi_{\mathrm{cls}}^u$ acts as a differentiable proxy 
for the indicator $\mathbf{1}\{\hat{\pi}(x_\nu') \neq \nu\}$, 
thereby encouraging $\hat{\pi}(x'_\nu) = \nu$. 
Such attacks are particularly powerful:  
(i) in classification, an adversary may compel predictions into sensitive categories 
of $\mc{Y}$; and  
(ii) the adversary may force deferral to a specific expert—possibly one 
known to perform poorly or to exhibit exploitable dependencies. 
For example, even if the clean input $x$ should defer to expert~1, 
a targeted perturbation may redirect the system to predict an unrelated class 
or to defer to a less reliable expert.

\begin{remark}
    In the two-stage case \citep{montreuil2025adversarial}, both targeted and untargeted attacks focus on the allocation policy, whereas in our setting adversaries can attack both the predictive and allocation policies.
\end{remark}

\section{DEFENDING AGAINST ATTACKS WITH GUARANTEES}

As previously discussed, current Learning-to-Defer approaches are highly sensitive to adversarial perturbations. This motivates the development of a formulation that explicitly incorporates adversarial robustness. Building on \citet{montreuil2025adversarial}, we generalize and extend the analysis to the one-stage L2D setting and introduce \emph{outcome-wise adversarial deferral losses} for both classification and regression. 

\subsection{Classification Setting}
\subsubsection{True Loss and Surrogates}
We now move from the one-stage attack setting to optimization objectives that enable learning a robust hypothesis $h \in \mc{H}$. To this end, we define an outcome-wise adversarial deferral loss by distinguishing perturbations according to their decision outcomes. This quantity upper-bounds the conventional single-perturbation adversarial deferral loss. 

Given an input $x$, let $x'_j \in B_p(x,\gamma)$ denote the adversarial example associated with outcome $j \in \mc{A}$, written as $x'_j = x + \delta_j$ where $\delta_j \in \mc{X}$ is the perturbation leading to outcome $j$. Collecting these perturbations, we write $\boldsymbol{\delta} = (\delta_1, \dots, \delta_{K+J})$ for the family of outcome-specific perturbations. For example, $x'_1$ denotes the adversarial input aligned with outcome $1 \in \mc{A}$.

\begin{restatable}[Outcome-wise Adversarial Deferral Loss for Classification]{lemma}{tlcrobustness}\label{lem:tlcrobustness}
For any hypothesis $h \in \mc{H}$, input–label pair $(x,y)$, 
and expert outputs $\mathbf{m} = (m_1,\dots,m_J)$, 
the \emph{outcome-wise adversarial deferral loss for classification} is
\begin{equation*}
    \widetilde{\ell}_{\mathrm{def}}^c\big(h, x, y, \mathbf{m}\big) 
    = \sum_{j=1}^{K+J} \mu_j(j, m_j, y)\mspace{-19mu}
      \sup_{x'_j \in B_p(x,\gamma)} \mspace{-19mu}
      \mathbf{1}\{\hat{h}(x'_j) = j\}.
\end{equation*}
\end{restatable}
Here, $\mu_j$ denotes a shifted cost. For $j \in \{1,\dots,K\}$, we set 
$\mu_j(j,y) = \alpha_j \mathbf{1}\{j \neq y\} + \beta_j$, which encodes the cost of predicting 
class $j$. For $j \in \{K+1,\dots,K+J\}$, we define $\mu_j(m_j,y) = c_{j-K}(m_{j-K},y)$, 
capturing the cost of deferring to expert $j$. 

We prove in Appendix \ref{proof:lem:tlcrobustness} that this quantity upper-bounds the conventional single-perturbation adversarial deferral loss, and therefore also upper-bounds the clean deferral loss in classification (Equation~\ref{class}). Since this formulation is NP-hard, we next introduce a tractable surrogate.

\begin{restatable}[Adversarial Surrogate Deferral Loss for Classification]{definition}{scrobustness}\label{lem:scrobustness}
For any hypothesis $h \in \mc{H}$, input–label pair $(x,y)$, and expert 
outputs $\mathbf{m} = (m_1,\dots,m_J)$, the \emph{adversarial surrogate deferral loss for 
classification} is defined as
\begin{equation*} \begin{aligned} \widetilde{\Phi}_{\mathrm{def}}^{c, u}(h&, x,y, \mathbf{m}) \mspace{-3mu}=\mspace{-8mu} \sum_{j=1}^{K+J}\mspace{-6mu} \sum_{i\not=j}\mspace{-3mu}\mu_i(i, m_i, y) \mspace{-24mu}\sup_{x'_j \in B_p(x,\gamma)}\mspace{-24mu} \Phi^{\rho, u}_{\mathrm{cls}}(h(x_j'),j). \end{aligned} \end{equation*}
\end{restatable}
Here, $\Phi^{\rho,u}_{\mathrm{cls}}(h(x'_j), j)$ denotes the margin-based surrogate loss evaluated 
on the adversarial input $x'_j$ associated with outcome $j \in \mc{A}$.

\subsubsection{Theoretical Guarantees} 

\paragraph{$\mc{H}$-consistency.}
The surrogate in Definition~\ref{lem:scrobustness} is a tractable relaxation of the outcome-wise adversarial loss in Lemma~\ref{lem:tlcrobustness}. 
A key requirement is that it be both Bayes- and $\mathcal{H}$-consistent with respect to the defined outcome-wise adversarial loss, 
ensuring that minimization recovers an optimal allocation policy for this loss. 
We establish these guarantees in the sense of \citet{Statistical, Steinwart2007HowTC}, showing that optimization over the surrogate yields asymptotically optimal policies for the proposed adversarial loss.

\begin{restatable}[$\mathcal{H}$-consistency bounds of $\widetilde{\Phi}^{c,u}_{\text{def}}$]{theorem}{consistency}
\label{theo:consistency}
Let $\mc{H}$ be symmetric and locally $\rho$-consistent. Then, for the set $\mc{A}$, any hypothesis $h \in \mc{H}$, and any distribution $\mc{D}$, the following holds:
\begin{equation*}
    \begin{aligned}
        & \mathcal{E}_{\widetilde{\ell}^c_{\text{def}}}(h) - \mathcal{E}_{\widetilde{\ell}^c_{\text{def}}}^B(\mc{H}) + \mathcal{U}_{\widetilde{\ell}^c_{\text{def}}}(\mc{H}) \\
        & \quad \leq \Psi^u(1) \Big( \mathcal{E}_{\widetilde{\Phi}^{c,u}_{\text{def}}}(h) - \mathcal{E}_{\widetilde{\Phi}^{c,u}_{\text{def}}}^\ast(\mc{H}) + \mathcal{U}_{\widetilde{\Phi}^{c,u}_{\text{def}}}(\mc{H}) \Big),
    \end{aligned}
\end{equation*}
with $\Psi^u(1) =
\begin{cases}
\log(2), & u=1, \\[0.3em]
\tfrac{1}{1-u}\big(2^{1-u}-1\big), & u \neq 1.
\end{cases}$
\end{restatable}

We prove this theorem using novel proof techniques, provided in Appendix~\ref{proof:theo:consistency}.  Intuitively, Theorem~\ref{theo:consistency} shows that the surrogate loss $\widetilde{\Phi}^{c,u}_{\text{def}}$ is $\mc{H}$-consistent up to the multiplicative constant $\Psi^u(1)$. If a sequence $(h_t)\subset\mc{H}$ satisfies 
\[
\mathcal{E}_{\widetilde{\Phi}^{c,u}_{\text{def}}}(h_t)-\mathcal{E}_{\widetilde{\Phi}^{c,u}_{\text{def}}}^\ast(\mc{H})+\mathcal{U}_{\widetilde{\Phi}^{c,u}_{\text{def}}}(\mc{H})\to 0,
\]
then by Theorem~\ref{theo:consistency} we have $\mathcal{E}_{\widetilde{\ell}^c_{\text{def}}}(h_t)-\mathcal{E}_{\widetilde{\ell}^c_{\text{def}}}^B(\mc{H})+\mathcal{U}_{\widetilde{\ell}^c_{\text{def}}}(\mc{H})\to 0$. Thus, minimizing the surrogate excess risk directly controls the excess risk of the defined outcome-wise adversarial deferral loss, ensuring that optimization with respect to $\widetilde{\Phi}^{c,u}_{\text{def}}$ converges to the Bayes-optimal policy for this loss. Finally, note that the minimizability gap vanishes for realizable distributions or when $\mc{H}$ is the set of all measurable functions $\mc{H}_{\mathrm{all}}$ \citep{Steinwart2007HowTC}.

\paragraph{Relaxing non-convexity.} While we established the consistency of the surrogate introduced in Definition~\ref{lem:scrobustness}, the formulation remains non-convex due to its reliance on the margin-based surrogate $\widetilde{\Phi}_{\text{cls}}^{\rho,u}$. 
Non-convexity complicates optimization and limits the practical applicability of the guarantee. 
To address this, we construct a smooth upper bound by replacing the margin loss with a tractable relaxation, following the principled approaches of \citet{Grounded, mao2023crossentropylossfunctionstheoretical, montreuil2025adversarial}. 
This relaxation preserves theoretical soundness while enabling tractable and stable optimization in practice.

\begin{restatable}[Smooth Adversarial Surrogate Losses]{lemma}{surrogatemulti} 
\label{lemma:surrogate_class}
Let $x \in \mc{X}$ be a clean input, and let $\rho > 0$ and $\kappa > 0$ 
be hyperparameters. The smooth adversarial surrogate losses are defined as
\begin{equation*} \begin{aligned} \widetilde{\Phi}_{\mathrm{cls, s}}^{u}(h, x,j) = &\, \Phi_{\mathrm{cls}}^u (h(x)/\rho, j) \\ & + \kappa \mspace{-15mu}\sup_{x'_j \in B_p(x,\gamma)} \mspace{-15mu} \big\| \overline{\Delta}_h(x'_j, j) - \overline{\Delta}_h(x, j) \big\|_2, \end{aligned} \end{equation*}
\end{restatable}
 We defer the proof in Appendix \ref{proof:lemma:surrogate_class}.  For any $x \in \mc{X}$, we define the pairwise margin differences as 
$\Delta_h(x, j, j') = h(x)_j - h(x)_{j'}$, 
and let $\overline{\Delta}_h(x, j) \in \mathbb{R}^{|\mc{A}|-1}$ 
be the vector of all pairwise differences, i.e., $\overline{\Delta}_h(x, j) = \big(\Delta_h(x, j, 1), \ldots, \Delta_h(x, j, j-1), \Delta_h(x, j, j+1), \ldots, \Delta_h(x, j, K+J)\big)$. The first term, $\Phi_{\mathrm{cls}}^u(h(x)/\rho, j)$, is the standard 
multiclass surrogate loss scaled by $\rho$, 
while the second term penalizes local instability by bounding the worst-case deviation 
of pairwise margins under adversarial perturbations. This yields a smooth tractable relaxation of the surrogate in Definition~\ref{lem:scrobustness}.

\begin{restatable}[Smooth Adversarial Surrogate Deferral Loss for Classification]{definition}{smoothscrobustness}\label{lem:smoothscrobustness}
For any hypothesis $h \in \mc{H}$, input–label pair $(x,y)$, and expert 
outputs $\mathbf{m} = (m_1,\dots,m_J)$, the \emph{smooth adversarial surrogate deferral loss for 
classification} is defined as
\begin{equation*} \begin{aligned} \widetilde{\Phi}_{\mathrm{def,s}}^{c, u}(h&, x,y, \mathbf{m}) \mspace{-3mu}=\mspace{-8mu} \sum_{j=1}^{K+J}\mspace{-6mu} \sum_{i\not=j}\mspace{-3mu}\mu_i(i, m_i, y) \mspace{-24mu}\sup_{x'_j \in B_p(x,\gamma)}\mspace{-24mu} \Phi^{u}_{\mathrm{cls,s}}(h(x_j'),j). \end{aligned} \end{equation*}
\end{restatable}
which yields the following pointwise upper bound.

\begin{restatable}[Pointwise Upper Bound for $\widetilde{\Phi}^{c,u}_{\mathrm{def,s}}$]{corollary}{smoothconsistency}
\label{theo:smooth-consistency}
For any hypothesis $h \in \mc{H}$, input--label pair $(x,y)$, and expert outputs $\mathbf{m} = (m_1,\dots,m_J)$, the following holds:
\begin{equation*}
    \widetilde{\Phi}^{c,u}_{\mathrm{def}}(h,x,y,\mathbf{m})
    \leq
    \widetilde{\Phi}^{c,u}_{\mathrm{def,s}}(h,x,y,\mathbf{m}).
\end{equation*}
\end{restatable}
Corollary~\ref{theo:smooth-consistency} shows that the smooth surrogate 
$\widetilde{\Phi}^{c,u}_{\mathrm{def,s}}$ pointwise upper-bounds 
$\widetilde{\Phi}^{c,u}_{\mathrm{def}}$. Combined with Theorem~\ref{theo:consistency}, this motivates optimizing the smooth relaxation as a tractable proxy for the nonsmooth adversarial surrogate.

\begin{definition}[RERM-C: Regularized ERM for $\widetilde{\Phi}^{c,u}_{\mathrm{def,s}}$] 
\label{erm_c_advl2d}
Assume $\mc{H}$ is symmetric and locally $\rho$-consistent. 
Let $\Omega: \mc{H} \to \mathbb{R}^+$ be a regularizer and let $\eta > 0$ be a hyperparameter. 
We define the regularized empirical risk minimization (ERM) objective as
\begin{equation*}
    \min_{h \in \mc{H}} \Bigg[
        \frac{1}{n} \sum_{k=1}^n 
        \widetilde{\Phi}^{c,u}_{\mathrm{def,s}}\big(h,x_k, y_k, \mathbf{m}_k\big) 
        + \eta \, \Omega(h)
    \Bigg],
\end{equation*}
\end{definition}

\subsection{Regression}
\subsubsection{True Loss and Surrogates}

Unlike classification, regression requires jointly learning both the main predictor $f \in \mc{F}$ and the allocation policy defined by the rejector $r \in \mc{R}$ (see Equation~\ref{regression}). 
This interdependence introduces additional complexity, making a direct extension of the classification analysis infeasible. 
To address this, we formalize an outcome-wise adversarial deferral loss for regression, which characterizes a worst-case upper bound under perturbations. 
This formulation extends the deferral framework to settings where both prediction and deferral are adversarially sensitive. 
The result is summarized in Lemma~\ref{lem:tlcrobustness-reg}, with the proof deferred to Appendix~\ref{proof:lem:tlcrobustness-reg}. 
\begin{restatable}[Outcome-wise Adversarial Deferral Loss for Regression]{lemma}{tlcrobustnessreg}
\label{lem:tlcrobustness-reg}
For any hypothesis $r \in \mc{R}$, predictor $f \in \mc{F}$, input–label pair $(x,t)$, 
and expert outputs $\mathbf{m} = (m_1,\dots,m_J)$, 
the \emph{outcome-wise adversarial deferral loss for regression} is defined as
\begin{equation*}
    \widetilde{\ell}_{\mathrm{def}}^r\big(r, f, t, x, \mathbf{m}\big) 
    = \sum_{j=1}^{J+1} \tilde{c}_j^r \mspace{-10mu}
      \sup_{x'_j \in B_p(x,\gamma)} \mspace{-10mu}
      \mathbf{1}\{\hat{r}(x'_j) = j\},
\end{equation*}
Here, $\tilde{c}_j^r = 
\begin{cases}
    \alpha_1\sup_{x'_1 \in B_p(x,\gamma)} L(f(x'_1), t)+\beta_1, & j = 1, \\[0.4em]
    c_j^r(m_{j-1}, t), & j >1.
\end{cases}$
\end{restatable}

To make the regression setting amenable to optimization, we construct a surrogate that corrects for the adaptive penalization introduced by adversarial perturbations.

\begin{restatable}[Adversarial Surrogate Deferral Loss for Regression]{definition}{srrobustness}
\label{lem:rcrobustness}
For any hypothesis $r \in \mc{R}$, predictor $f \in \mc{F}$, input–label pair $(x,t)$, 
and expert outputs $\mathbf{m} = (m_1,\dots,m_J)$, 
the \emph{adversarial surrogate deferral loss for regression} is defined as
\begin{equation*} \begin{aligned} \widetilde{\Phi}_{\mathrm{def}}^{r, u}(r&,f,x, t, \mathbf{m}) = -(J-1)\tilde{c}_1^r(f,x,t) \\ & + \sum_{j=1}^{J+1}\mspace{-6mu} \sum_{i\not=j}\tilde{c}_i^r(f,x,m_{i-1},t) \mspace{-24mu}\sup_{x'_j \in B_p(x,\gamma)}\mspace{-24mu} \Phi^{\rho, u}_{\mathrm{cls}}(r(x_j'),j). \end{aligned} \end{equation*}
\end{restatable}
This surrogate loss provides a tractable relaxation of the outcome-wise adversarial deferral loss, 
while explicitly incorporating adaptive penalization through $\tilde{c}_1^r$.

\begin{remark}
    This surrogate differs fundamentally from both its classification counterpart and the two-stage setting \citep{montreuil2025adversarial}, owing to the explicit involvement of the learnable predictor $f \in \mc{F}$.
\end{remark}

\subsubsection{Theoretical Guarantees} 
Similarly to the classification setting, we introduced both an adversarial surrogate deferral loss and its outcome-wise counterpart. It is therefore necessary to establish that the surrogate in Definition~\ref{lem:rcrobustness} is Bayes- and $(\mc{R}, \mc{F})$-consistent. We prove the following theorem in Appendix \ref{proof:theo:r_consistency}. 

\begin{restatable}[$(\mc{R}, \mc{F})$-consistency bounds of $\widetilde{\Phi}^{r,u}_{\text{def}}$]{theorem}{rconsistency}
\label{theo:r_consistency}
Let $\mc{R}$ be symmetric and locally $\rho$-consistent. Then, for the set $\mc{A}^{r}$, any hypothesis $r \in \mc{R}$ and $f \in \mc{F}$, and any distribution $\mc{D}$, the following holds:
\begin{equation*}
    \begin{aligned}
        & \mathcal{E}_{\widetilde{\ell}^r_{\text{def}}}(r,f) - \mathcal{E}_{\widetilde{\ell}^r_{\text{def}}}^B(\mc{R}, \mc{F}) + \mathcal{U}_{\widetilde{\ell}^r_{\text{def}}}(\mc{R}, \mc{F}) \\
        & \leq \overline{\Gamma}^u(1) \Big( \mathcal{E}_{\widetilde{\Phi}^{r,u}_{\text{def}}}(r,f) - \mathcal{E}_{\widetilde{\Phi}^{r,u}_{\text{def}}}^\ast(\mc{R}, \mc{F}) + \mathcal{U}_{\widetilde{\Phi}^{r,u}_{\text{def}}}(\mc{R}, \mc{F}) \Big),
    \end{aligned}
\end{equation*}
with $\overline{\Gamma}^u(1) = \max\{1, \Psi^u(1)\}$.
\end{restatable}
Theorem~\ref{theo:r_consistency} shows that the adversarial surrogate loss for regression is $(\mc{R}, \mc{F})$-consistent up to the factor $\max\{1, \Psi^u(1)\}$ with respect to the defined outcome-wise adversarial loss. 
Importantly, this factor differs from the classification case (see Theorem~\ref{theo:consistency}) precisely because regression involves the learnable predictor $f \in \mc{F}$ in addition to the rejector $r \in \mc{R}$. 
As a result, the analysis must bound a joint function of $(r,f)$ rather than a single function, which makes the regression case fundamentally more challenging. Indeed, if two sequences $(f_t) \subset \mc{F}$ and $(r_t) \subset \mc{R}$ satisfy $\mathcal{E}_{\widetilde{\Phi}^{r,u}_{\text{def}}}(r_t,f_t) - \mathcal{E}_{\widetilde{\Phi}^{r,u}_{\text{def}}}^\ast(\mc{R}, \mc{F}) + \mathcal{U}_{\widetilde{\Phi}^{r,u}_{\text{def}}}(\mc{R}, \mc{F}) \rightarrow 0$, then by Theorem \ref{theo:r_consistency} we have $\mathcal{E}_{\widetilde{\ell}^r_{\text{def}}}(r_t,f_t) - \mathcal{E}_{\widetilde{\ell}^r_{\text{def}}}^B(\mc{R}, \mc{F}) + \mathcal{U}_{\widetilde{\ell}^r_{\text{def}}}(\mc{R}, \mc{F}) \rightarrow 0$. Therefore, minimizing the surrogate recovers an optimal predictor-rejector pair for the defined adversarial loss. 
In this case, the minimizability gap vanishes under realizability with respect to the pair $(r,f)$, provided that $\mc{F}=\mc{F}_{\text{all}}$ and $\mc{R}=\mc{R}_{\text{all}}$.

Analogous to the classification case, we apply Lemma~\ref{lemma:surrogate_class} to upper-bound the surrogate defined in Definition~\ref{lem:rcrobustness} by a smooth relaxation. 
\begin{restatable}[Smooth Adversarial Surrogate Deferral Loss for Regression]{corollary}{smoothsrrobustness}
\label{lem:smoothrcrobustness}
For any hypothesis $r \in \mc{R}$, predictor $f \in \mc{F}$, input–label pair $(x,t)$, 
and expert outputs $\mathbf{m} = (m_1,\dots,m_J)$, 
the \emph{smooth adversarial surrogate deferral loss for regression} is defined as
\begin{equation*}
    \begin{aligned}
        \widetilde{\Phi}^{r,u}_{\mathrm{def,s}}(r, &f,  x,t,\mathbf{m})  =  -(J-1)\tilde{c}_1^r(f(x),t) \\
        &+ \sum_{j=1}^{J+1} \sum_{i\not=j}\tilde{c}_i^r(f,x,m_{i-1},t) \widetilde{\Phi}^{u}_{\mathrm{cls, s}}(r(x),j).
    \end{aligned}
\end{equation*}
\end{restatable}
This yields a novel surrogate with guarantees while ensuring tractable optimization.

\begin{restatable}[Pointwise Upper Bound for $\widetilde{\Phi}^{r,u}_{\text{def, s}}$]{corollary}{rconsistencycor}
\label{corr:r_consistency}
For any hypothesis $r \in \mc{R}$, predictor $f \in \mc{F}$, input--label pair $(x,t)$, and expert outputs $\mathbf{m} = (m_1,\dots,m_J)$, the following holds:
\begin{equation*}
    \widetilde{\Phi}^{r,u}_{\mathrm{def}}(r,f,x,t,\mathbf{m})
    \leq
    \widetilde{\Phi}^{r,u}_{\mathrm{def,s}}(r,f,x,t,\mathbf{m}).
\end{equation*}
\end{restatable}
Corollary~\ref{corr:r_consistency} shows that the smooth surrogate $\widetilde{\Phi}^{r,u}_{\mathrm{def,s}}$ pointwise upper-bounds $\widetilde{\Phi}^{r,u}_{\mathrm{def}}$. This provides a tractable optimization objective aligned with the adversarial surrogate analyzed in Theorem~\ref{theo:r_consistency}. 

Hence, it motivates the introduction of a regularized ERM algorithm for regression (RERM-R).

\begin{definition}[RERM-R: Regularized ERM for $\widetilde{\Phi}^{r,u}_{\mathrm{def,s}}$] 
\label{erm_r_advl2d}
Assume $\mc{R}$ is symmetric and locally $\rho$-consistent. 
Let $\Omega: \mc{R} \times \mc{F} \to \mathbb{R}_{+}$ be a convex regularizer, and let $\eta > 0$ be a hyperparameter. 
The regularized ERM objective is
\begin{equation*}
    \min_{\,r \in \mc{R},\, f \in \mc{F}} 
    \mspace{-5mu}\Bigg[
        \frac{1}{n} \sum_{k=1}^n 
        \widetilde{\Phi}^{r,u}_{\mathrm{def,s}}\!\big(r, f,x_k, t_k, \mathbf{m}_k\big) 
        + \eta \,\Omega(r,f)
    \Bigg]
\end{equation*}
\end{definition}

We will use this algorithm in Subsection~\ref{reg_task}. Its per-epoch computational cost in the \(h\)-score setting is summarized below.

\begin{proposition}[Epoch cost of RERM-C in the \(h\)-score setting]
\label{prop:epoch-cost-h_label}
Consider \(n\) training examples processed in mini-batches of size \(B\), and let \(\mathcal A\) denote the action space (for instance, \(\mathcal A=\{1,\dots,K+J\}\)). Suppose that each inner maximization is carried out by \(\mathrm{PGD}(T)\), that is, by \(T\) projected-gradient steps. Let \(C_{\mathrm{fwd}}\) and \(C_{\mathrm{bwd}}\) denote the costs of one forward and one backward pass through the score network \(h\), respectively. Then one epoch of RERM-C has computational cost
\begin{equation*}
\label{eq:epoch-cost_label}
n\bigl(1+|\mathcal A|T\bigr)\bigl(C_{\mathrm{fwd}}+C_{\mathrm{bwd}}\bigr).
\end{equation*}
Moreover, the peak memory requirement is the same as that of a standard forward-backward pass, up to the additional storage needed for one adversarial copy of each input currently being optimized.
\end{proposition}

% below are written by Letian, please review it before submission.
\section{EXPERIMENTS}
We evaluate the robustness of one-stage L2D under adversarial perturbations, focusing on both classification 
and regression tasks. Our experiments show that standard one-stage L2D baselines can degrade sharply under adversarial perturbations, with attacks affecting both predictive accuracy and deferral decisions. By contrast, our algorithms (RERM-C and RERM-R) improve attacked performance while maintaining competitive clean performance. All results are averaged over three independent runs. 

\phantomsection
\paragraph*{Comparison.}
\citet{montreuil2025adversarial} proposes a defense mechanism tailored to a more restrictive \emph{two-stage} setting, in which only the router is learned while the experts are fixed. As a result, a direct comparison is not meaningful, since the learning setup and objectives differ fundamentally from ours. We therefore focus on standard one-stage L2D baselines; adapting generic adversarial training methods developed for pure classification problems \citep{awasthi2021calibrationconsistencyadversarialsurrogate} to the augmented L2D action space is beyond the scope of this work.

\phantomsection
\paragraph*{Metrics.} 
We report the following evaluation metrics: 
\textit{Clean Accuracy} (C.Acc, \%) — the overall accuracy of the L2D system, accounting for both model predictions and expert deferrals on clean inputs; 
\textit{Untargeted Accuracy} (U.Acc, \%) — system accuracy under untargeted adversarial attack; 
\textit{Targeted Accuracy} (T.Acc, \%) — system accuracy under a targeted attack toward a randomly selected target action; 
\textit{Adversarial Deferral Loss} (Def.Loss) — the empirical outcome-wise adversarial deferral loss.

\subsection{Classification Setting} \label{class_setting}
We evaluate on CIFAR-10 \citep{krizhevsky2009learning} and 
DermaMNIST from the MedMNIST benchmark suite \citep{medmnistv1, medmnistv2, dermamnist1, dermamnist2}. 
As a baseline, we compare against the consistent deferral framework of \citet{mozannar2021consistent, mao2024principledapproacheslearningdefer}, 
which represent the standard approach for one-stage L2D.  In particular, we instantiate our method with the logistic loss ($u=1$) for RERM-C, 
as introduced in Definition~\ref{erm_c_advl2d}. The main paper reports results on CIFAR-10, while additional results on DermaMNIST are deferred to the Appendix.

\subsubsection{CIFAR10}

\phantomsection
\paragraph*{Setting.}
The augmented classifier is implemented using ResNet-4 \citep{he2015deepresiduallearningimage} and trained with the AdamW optimizer \citep{kingma2017adammethodstochasticoptimization} for 400 epochs. As experts, we employ three ResNet-16 models, each trained on a subset of the dataset; their accuracies are reported in Appendix~\ref{cifar10_exp}. The consultation costs are set as follows: $\beta_{j \leq K}=0$ for predictions, and $\beta_{K+1}=0.05$, $\beta_{K+2}=0.075$, and $\beta_{K+3}=0.1$ for the experts. The baseline method uses a learning rate of $0.005$, while our approach employs a learning rate of $0.01$. For the PGD attack, we set $\epsilon=0.03137$, and in our method we additionally use the hyperparameters $\rho=1.0$ and $\nu=0.002$.

\begin{table}[ht]
  \centering
  \resizebox{\columnwidth}{!}{%
    \begin{tabular}{@{}lcccccc@{}}
      \toprule
        & C.Acc & U.Acc & T.Acc & Def.Loss \\
      \midrule
      \citet{mao2024principledapproacheslearningdefer} 
        &  $82.67 \pm 2.06$ 
        & $27.47 \pm 0.63$ 
        & $19.80 \pm 0.48$ 
        & $0.75 \pm 0.02$  \\
      Ours 
        &  $75.60 \pm 1.87$ 
        & $52.00 \pm 1.31$ 
        & $68.67 \pm 1.70$ 
        & $0.52 \pm 0.01$ \\
      \bottomrule
    \end{tabular}%
  }
\caption{Performance under clean and adversarial inputs, compared against \citet{mao2024principledapproacheslearningdefer}.}
  \label{tab:cifar10}
\end{table}

\phantomsection
\paragraph*{Results.} 
Table~\ref{tab:cifar10} reports performance on the CIFAR-10 dataset, comparing our approach with \citet{mao2024principledapproacheslearningdefer}. 
The baseline achieves slightly higher clean accuracy ($82.67$ vs.\ $75.60$), but this advantage vanishes under adversarial perturbations: its accuracy collapses to $27.47$ (untargeted) and $19.80$ (targeted). 
In contrast, our method preserves robustness, improving untargeted accuracy by more than $24\%$ and targeted accuracy by nearly $49\%$. 
Furthermore, our approach attains a substantially lower empirical adversarial deferral loss under the outcome-wise evaluation metric of Lemma~\ref{lem:tlcrobustness}. 
These results indicate that our surrogate-based methods improve attacked performance while retaining competitive clean performance.

\subsubsection{DermaMNIST}\label{dermnist}

\phantomsection
\paragraph*{Setting.} 
DermaMNIST is a subset of the MedMNIST dataset consisting of biomedical images for 7-class classification. 
The augmented classifier is implemented using ResNet-18 and trained for 100 epochs. 
As experts, we construct three specialized classifiers, each responsible for a randomly assigned subset of three classes (with overlap), predicting correctly with probability $p=0.85$ on their assigned classes and uniformly at random otherwise; their accuracies are reported in the Appendix. 
The consultation costs are set as $\beta_{j \leq K}=0$ for predictions, and $\beta_{K+1}=0.05$, $\beta_{K+2}=0.075$, and $\beta_{K+3}=0.125$ for the experts. 
Both the baseline method and our approach use a learning rate of $0.005$. 
For the PGD attack, we set $\epsilon=0.03137$, while our approach additionally uses the hyperparameters $\rho=1.75$ and $\nu=0.001$.

\phantomsection
\paragraph*{Results.}
\begin{table}[ht]
  \centering
    \resizebox{\columnwidth}{!}{\begin{tabular}{@{}lcccccc@{}}
      \toprule
       & C.Acc & U.Acc & T.Acc & Def.Loss  \\
      \midrule
      \citet{mao2024principledapproacheslearningdefer}  & $83.39$ & $30.82$ & $27.08$ & $69.60$  \\
      Ours &  $81.80$ & $71.12$ & $80.65$ & $31.66$  \\
      \bottomrule
    \end{tabular}%
    }
  \caption{Performance under clean and adversarial inputs, compared against the approach of \citet{mao2024principledapproacheslearningdefer}.}
  \label{tab:dermamnist}
\end{table}

On the DermaMNIST dataset, Table~\ref{tab:dermamnist} shows that our approach remains close to the baseline under the clean setting. Under adversarial perturbations, our approach preserves substantially higher targeted and untargeted accuracy, while also achieving a lower empirical adversarial deferral loss.

\subsection{Regression Task}\label{reg_task}
We evaluate our approach on the Communities and Crime dataset \citep{communities_and_crime_183} and the Insurance Company Benchmark (COIL 2000) \citep{insurance_company_benchmark_(coil_2000)_125}.  As a baseline, we compare against \citet{mao2024regressionmultiexpertdeferral}, which represents the standard one-stage L2D approach. 
For our method, we instantiate the logistic loss ($u=1$) within RERM-R, as introduced in Definition~\ref{erm_r_advl2d}. The main paper reports results on Communities, while additional results on COIL 2000 are deferred to the Appendix.

\subsubsection{Communities and Crime}

\phantomsection
\paragraph*{Setting.} 
The rejector is implemented as an MLP and trained using the AdamW optimizer \citep{kingma2017adammethodstochasticoptimization} for 500 epochs. 
The main predictor is a linear layer. 
As experts, we employ three MLPs specialized in different socio-economic factors (e.g., demographics, economics, housing), and report their accuracies in the Appendix. 
The consultation costs are set as $\beta_{1}=0$ for the main predictor, and $\beta_{2}=0.04$, $\beta_{3}=0.05$, and $\beta_{4}=0.07$ for experts $m_1, m_2, m_3$. 
Both the baseline and our method are trained with AdamW; the baseline uses a learning rate of $0.005$, while our approach employs $0.01$. 
For the PGD attack, we set $\epsilon=0.5$; our method additionally uses the hyperparameters $\rho=2.5$ and $\nu=0.005$.

\begin{table}[ht]
  \centering
  \resizebox{\columnwidth}{!}{%
    \begin{tabular}{@{}lcccccc@{}}
      \toprule
        & C.Acc & U.Acc & T.Acc & Def.Loss \\
      \midrule
      \citet{mao2024regressionmultiexpertdeferral}  
        & $9.96 \pm 0.21$ 
        & $18.08 \pm 0.44$ 
        & $56.74 \pm 1.57$ 
        & $18.93 \pm 0.40$  \\
      Ours 
        & $12.09 \pm 0.39$ 
        & $12.13 \pm 0.31$ 
        & $19.67 \pm 0.12$ 
        & $12.94 \pm 0.26$  \\
      \bottomrule
    \end{tabular}%
  }
  \caption{Performance under clean and adversarial inputs, compared with \citet{mao2024regressionmultiexpertdeferral}. All accuracies are reported as RMSE, where lower values indicate better performance. }
  \label{tab:comm-cirme}
\end{table}

\phantomsection
\paragraph*{Results.} 
Table~\ref{tab:comm-cirme} compares our method with \citet{mao2024regressionmultiexpertdeferral} on the Communities \& Crime dataset. 
All accuracies are reported as RMSE, where lower values indicate higher accuracy. 
On clean inputs, our approach performs comparably to the baseline. 
Under adversarial perturbations, however, our method provides substantial gains: in the targeted setting, RMSE is reduced by nearly $37\%$, and in the untargeted setting, it remains consistently lower than the baseline. 
Equally important, our approach achieves a markedly lower deferral loss. 
This gap suggests that our method attains a more favorable trade-off between predictive error, robustness, and consultation cost under the reported evaluation pipeline.

\subsubsection{Insurance Company Benchmark (COIL 2000)}

\phantomsection
\paragraph*{Setting.}

The rejector is implemented using an MLP and trained with the AdamW optimizer \citep{kingma2017adammethodstochasticoptimization} for 25 epochs. The main predictor is a linear layer. As experts, we employ four regression MLPs, each focusing on different customer segments (demographics, product ownership, high-value customers) and generating predictions using rules and noise; their accuracies are reported in the Appendix. The consultation costs are set as follows: $\beta_{1}=0$ for the main predictor, and $\beta_2=0.035$, $\beta_3=0.04$, $\beta_4=0.045$ and $\beta_5=0.05$ for the experts. The baseline method uses a learning rate of $0.005$, while our approach employs a learning rate of $0.01$. For the PGD attack, we set $\epsilon=2$, and in our method we additionally use the hyperparameters $\rho=2.75$ and $\nu=0.01$.

\phantomsection
\paragraph*{Results.}

\begin{table}[ht]
  \centering\resizebox{\columnwidth}{!}{%
    \begin{tabular}{@{}lcccccc@{}}
      \toprule
      & C.Acc & U.Acc & T.Acc & Def.Loss \\
      \midrule
      \citet{mao2024regressionmultiexpertdeferral} &  $7.02$ & $11.61$ & $8.31$ & $11.98$  \\
      Ours &  $7.39$ & $7.41$ & $7.40$ & $7.81$  \\
      \bottomrule
    \end{tabular}}%
  \caption{Performance under clean and adversarial inputs, compared against the approach of \citet{mao2024regressionmultiexpertdeferral}.}
  \label{tab:coil}
\end{table}

On the COIL-2000 dataset, our approach remains close to the baseline under the clean setting while improving both attacked RMSE and adversarial deferral loss. This suggests that the approach transfers reasonably well across the reported regression benchmarks.

\section{CONCLUSION}
We presented a framework for adversarial robustness in one-stage Learning-to-Defer (L2D), addressing both classification and regression tasks. Our work makes three key advances: we formalized untargeted and targeted attacks that reveal how adversaries can jointly exploit prediction and deferral; 
we proposed outcome-wise adversarial surrogate losses with tractable relaxations and established consistency guarantees for the proposed adversarial losses; 
and we demonstrated empirically, across diverse benchmarks, that our methods improve attacked performance while maintaining competitive clean accuracy. 

\section*{ACKNOWLEDGMENTS}
This research was supported by the National Research Foundation, Singapore, under its AI Singapore Programme (AISG Award No.\ AISG2-PhD-2023-01-041-J) and by A*STAR, and forms part of the DesCartes programme, which is supported by the National Research Foundation, Prime Minister's Office, Singapore, under its Campus for Research Excellence and Technological Enterprise (CREATE) programme.

\bibliographystyle{plainnat}
\bibliography{biblio.bib}

@inproceedings{
montreuil2025askaskktwostage,
title={Why Ask One When You Can Ask \$k\$? Learning-to-Defer to the Top-\$k\$ Experts},
author={Yannis Montreuil and Axel Carlier and Lai Xing Ng and Wei Tsang Ooi},
booktitle={The Fourteenth International Conference on Learning Representations},
year={2026},
url={https://openreview.net/forum?id=mGbEv4kVoG}
}

@inproceedings{Keswani,
author = {Keswani, Vijay and Lease, Matthew and Kenthapadi, Krishnaram},
title = {Towards Unbiased and Accurate Deferral to Multiple Experts},
year = {2021},
isbn = {9781450384735},
publisher = {Association for Computing Machinery},
address = {New York, NY, USA},
url = {https://doi.org/10.1145/3461702.3462516},
doi = {10.1145/3461702.3462516},
booktitle = {Proceedings of the 2021 AAAI/ACM Conference on AI, Ethics, and Society},
pages = {154–165},
numpages = {12},
location = {Virtual Event, USA},
series = {AIES '21}
}

@inproceedings{Kerrigan,
 author = {Kerrigan, Gavin and Smyth, Padhraic and Steyvers, Mark},
 booktitle = {Advances in Neural Information Processing Systems},
 editor = {M. Ranzato and A. Beygelzimer and Y. Dauphin and P.S. Liang and J. Wortman Vaughan},
 pages = {4421--4434},
 publisher = {Curran Associates, Inc.},
 title = {Combining Human Predictions with Model Probabilities via Confusion Matrices and Calibration},
 url = {https://proceedings.neurips.cc/paper_files/paper/2021/file/234b941e88b755b7a72a1c1dd5022f30-Paper.pdf},
 volume = {34},
 year = {2021}
}

@inproceedings{liu2024mitigating,
  title={Mitigating Underfitting in Learning to Defer with Consistent Losses},
  author={Liu, Shuqi and Cao, Yuzhou and Zhang, Qiaozhen and Feng, Lei and An, Bo},
  booktitle={International Conference on Artificial Intelligence and Statistics},
  pages={4816--4824},
  year={2024},
  organization={PMLR}
}

@InProceedings{Tailor,
  title = 	 {Learning to Defer to a Population: A Meta-Learning Approach},
  author =       {Tailor, Dharmesh and Patra, Aditya and Verma, Rajeev and Manggala, Putra and Nalisnick, Eric},
  booktitle = 	 {Proceedings of The 27th International Conference on Artificial Intelligence and Statistics},
  pages = 	 {3475--3483},
  year = 	 {2024},
  editor = 	 {Dasgupta, Sanjoy and Mandt, Stephan and Li, Yingzhen},
  volume = 	 {238},
  series = 	 {Proceedings of Machine Learning Research},
  month = 	 {02--04 May},
  publisher =    {PMLR},
  url = 	 {https://proceedings.mlr.press/v238/tailor24a.html}
}

@inproceedings{Benz,
  title={Counterfactual inference of second opinions},
  author={Benz, Nina L Corvelo and Rodriguez, Manuel Gomez},
  booktitle={Uncertainty in Artificial Intelligence},
  pages={453--463},
  year={2022},
  organization={PMLR}
}

@inproceedings{Hemmer,
  title     = {Forming Effective Human-{AI} Teams: Building Machine Learning Models that Complement the Capabilities of Multiple Experts},
  author    = {Hemmer, Patrick and Schellhammer, Sebastian and Vössing, Michael and Jakubik, Johannes and Satzger, Gerhard},
  booktitle = {Proceedings of the Thirty-First International Joint Conference on
               Artificial Intelligence, {IJCAI-22}},
  publisher = {International Joint Conferences on Artificial Intelligence Organization},
  editor    = {Lud De Raedt},
  pages     = {2478--2484},
  year      = {2022},
  month     = {7},
  note      = {Main Track},
  doi       = {10.24963/ijcai.2022/344},
  url       = {https://doi.org/10.24963/ijcai.2022/344},
}

@techreport{krizhevsky2009learning,
  title={Learning Multiple Layers of Features from Tiny Images},
  author={Krizhevsky, Alex},
  year={2009},
  institution={University of Toronto},
  note={Technical report}
}

@ARTICLE{Ohn_Aldrich1997-wn,
  title   = "Fisher and the making of maximum likelihood 1912-1922",
  author  = "Ohn Aldrich, R A",
  journal = "Statistical Science",
  volume  =  12,
  number  =  3,
  pages   = "162--179",
  year    =  1997
}

@article{kingma2017adammethodstochasticoptimization,
  title={Adam: A method for stochastic optimization},
  author={Kingma, Diederik P and Ba, Jimmy},
  journal={arXiv preprint arXiv:1412.6980},
  year={2014}
}

@inproceedings{Ghosh,
author = {Ghosh, Aritra and Kumar, Himanshu and Sastry, P. S.},
title = {Robust loss functions under label noise for deep neural networks},
year = {2017},
publisher = {AAAI Press},
booktitle = {Proceedings of the Thirty-First AAAI Conference on Artificial Intelligence},
pages = {1919–1925},
numpages = {7},
location = {San Francisco, California, USA},
series = {AAAI'17}
}

@inproceedings{Cao_Mozannar_Feng_Wei_An_2023,
author = {Cao, Yuzhou and Mozannar, Hussein and Feng, Lei and Wei, Hongxin and An, Bo},
title = {In defense of softmax parametrization for calibrated and consistent learning to defer},
year = {2024},
publisher = {Curran Associates Inc.},
address = {Red Hook, NY, USA},
booktitle = {Proceedings of the 37th International Conference on Neural Information Processing Systems},
articleno = {1671},
numpages = {19},
location = {New Orleans, LA, USA},
series = {NIPS '23}
}

@article{
joshi2021learning,
title={Learning-to-defer for sequential medical decision-making under uncertainty},
author={Shalmali Joshi and Sonali Parbhoo and Finale Doshi-Velez},
journal={Transactions on Machine Learning Research},
issn={2835-8856},
year={2023},
url={https://openreview.net/forum?id=0pn3KnbH5F},
note={}
}

@inproceedings{strong2025trustworthy,
  title={Trustworthy and Practical AI for Healthcare: A Guided Deferral System with Large Language Models},
  author={Strong, Joshua and Men, Qianhui and Noble, J Alison},
  booktitle={Proceedings of the AAAI Conference on Artificial Intelligence},
  volume={39},
  pages={28413--28421},
  year={2025}
}

@inproceedings{
palomba2025a,
title={A Causal Framework for Evaluating Deferring Systems},
author={Filippo Palomba and Andrea Pugnana and Jose Manuel Alvarez and Salvatore Ruggieri},
booktitle={The 28th International Conference on Artificial Intelligence and Statistics},
year={2025},
url={https://openreview.net/forum?id=mkkFubLdNW}
}

@inproceedings{Narasimhan,
 author = {Narasimhan, Harikrishna and Jitkrittum, Wittawat and Menon, Aditya K and Rawat, Ankit and Kumar, Sanjiv},
 booktitle = {Advances in Neural Information Processing Systems},
 editor = {S. Koyejo and S. Mohamed and A. Agarwal and D. Belgrave and K. Cho and A. Oh},
 pages = {29292--29304},
 publisher = {Curran Associates, Inc.},
 title = {Post-hoc estimators for learning to defer to an expert},
 url = {https://proceedings.neurips.cc/paper_files/paper/2022/file/bc8f76d9caadd48f77025b1c889d2e2d-Paper-Conference.pdf},
 volume = {35},
 year = {2022}
}

@inproceedings{
montreuil2024twostagelearningtodefermultitasklearning,
title={A Two-Stage Learning-to-Defer Approach for Multi-Task Learning},
author={Yannis Montreuil and Yeo Shu Heng and Axel Carlier and Lai Xing Ng and Wei Tsang Ooi},
booktitle={Forty-second International Conference on Machine Learning},
year={2025},
url={https://openreview.net/forum?id=qmeNQpLiG5}
}

@article{goodfellow2014explaining,
  title={Explaining and harnessing adversarial examples},
  author={Goodfellow, Ian J and Shlens, Jonathon and Szegedy, Christian},
  journal={arXiv preprint arXiv:1412.6572},
  year={2014}
}

@InProceedings{verma_multiple,
  title = 	 {Learning to Defer to Multiple Experts: Consistent Surrogate Losses, Confidence Calibration, and Conformal Ensembles},
  author =       {Verma, Rajeev and Barrejon, Daniel and Nalisnick, Eric},
  booktitle = 	 {Proceedings of The 26th International Conference on Artificial Intelligence and Statistics},
  pages = 	 {11415--11434},
  year = 	 {2023},
  editor = 	 {Ruiz, Francisco and Dy, Jennifer and van de Meent, Jan-Willem},
  volume = 	 {206},
  series = 	 {Proceedings of Machine Learning Research},
  month = 	 {25--27 Apr},
  publisher =    {PMLR},
  url = 	 {https://proceedings.mlr.press/v206/verma23a.html}
}

@inproceedings{Verma2022LearningTD,
  title={Learning to Defer to Multiple Experts: Consistent Surrogate Losses, Confidence Calibration, and Conformal Ensembles},
  author={Rajeev Verma and Daniel Barrejon and Eric Nalisnick},
  booktitle={International Conference on Artificial Intelligence and Statistics},
  year={2022},
  url={https://api.semanticscholar.org/CorpusID:253237048}
}

@inproceedings{mozannar2021consistent,
author = {Mozannar, Hussein and Sontag, David},
title = {Consistent estimators for learning to defer to an expert},
year = {2020},
publisher = {JMLR.org},
booktitle = {Proceedings of the 37th International Conference on Machine Learning},
articleno = {656},
numpages = {12},
series = {ICML'20}
}

@article{Gowal2020UncoveringTL,
  title={Uncovering the Limits of Adversarial Training against Norm-Bounded Adversarial Examples},
  author={Sven Gowal and Chongli Qin and Jonathan Uesato and Timothy A. Mann and Pushmeet Kohli},
  journal={ArXiv},
  year={2020},
  volume={abs/2010.03593},
  url={https://api.semanticscholar.org/CorpusID:222208628}
}

@article{Madry2017TowardsDL,
  title={Towards Deep Learning Models Resistant to Adversarial Attacks},
  author={Aleksander Madry and Aleksandar Makelov and Ludwig Schmidt and Dimitris Tsipras and Adrian Vladu},
  journal={ArXiv},
  year={2017},
  volume={abs/1706.06083},
  url={https://api.semanticscholar.org/CorpusID:3488815}
}

@article{Steinwart2007HowTC,
  title={How to Compare Different Loss Functions and Their Risks},
  author={Ingo Steinwart},
  journal={Constructive Approximation},
  year={2007},
  volume={26},
  pages={225-287},
  url={https://api.semanticscholar.org/CorpusID:16660598}
}

@inproceedings{he2015deepresiduallearningimage,
  title={Deep residual learning for image recognition},
  author={He, Kaiming and Zhang, Xiangyu and Ren, Shaoqing and Sun, Jian},
  booktitle={Proceedings of the IEEE conference on computer vision and pattern recognition},
  pages={770--778},
  year={2016}
}

@article{zhang2018generalizedcrossentropyloss,
  title={Generalized cross entropy loss for training deep neural networks with noisy labels},
  author={Zhang, Zhilu and Sabuncu, Mert},
  journal={Advances in neural information processing systems},
  volume={31},
  year={2018}
}

@article{jia2017adversarial,
  title={Adversarial Examples for Evaluating Reading Comprehension Systems},
  author={Jia, Robin and Liang, Percy},
  journal={arXiv preprint arXiv:1707.07328},
  year={2017}
}

@article{bartlett1,
author = {Bartlett, Peter and Jordan, Michael and McAuliffe, Jon},
year = {2006},
month = {02},
pages = {138-156},
title = {Convexity, Classification, and Risk Bounds},
volume = {101},
journal = {Journal of the American Statistical Association},
doi = {10.1198/016214505000000907}
}

@article{madras2018predict,
  title={Predict responsibly: improving fairness and accuracy by learning to defer},
  author={Madras, David and Pitassi, Toni and Zemel, Richard},
  journal={Advances in neural information processing systems},
  volume={31},
  year={2018}
}

@article{chenfrugalgpt,
  title={FrugalGPT: How to Use Large Language Models While Reducing Cost and Improving Performance},
  author={Chen, Lingjiao and Zaharia, Matei and Zou, James},
  journal={Transactions on Machine Learning Research},
    year={2024}
}

@inproceedings{
jitkrittum2025universal,
title={Universal Model Routing for Efficient {LLM} Inference},
author={Wittawat Jitkrittum and Harikrishna Narasimhan and Ankit Singh Rawat and Jeevesh Juneja and Congchao Wang and Zifeng Wang and Alec Go and Chen-Yu Lee and Pradeep Shenoy and Rina Panigrahy and Aditya Krishna Menon and Sanjiv Kumar},
booktitle={The Fourteenth International Conference on Learning Representations},
year={2026},
url={https://openreview.net/forum?id=ka82fvJ5f1}
}

@inproceedings{
montreuil2025optimalqueryallocationextractive,
title={Optimal Query Allocation in Extractive {QA} with {LLM}s: A Learning-to-Defer Framework with Theoretical Guarantees},
author={Yannis Montreuil and Yeo Shu Heng and Axel Carlier and Lai Xing Ng and Wei Tsang Ooi},
booktitle={The 29th International Conference on Artificial Intelligence and Statistics},
year={2026},
url={https://openreview.net/forum?id=kEVupwepTq}
}

@inproceedings{
montreuil2025adversarial,
title={Adversarial Robustness in Two-Stage Learning-to-Defer: Algorithms and Guarantees},
author={Yannis Montreuil and Axel Carlier and Lai Xing Ng and Wei Tsang Ooi},
booktitle={Forty-second International Conference on Machine Learning},
year={2025},
url={https://openreview.net/forum?id=h3KHwZCnxH}
}

@inproceedings{
montreuil2026online,
title={Online Learning-to-Defer with Varying Experts},
author={Yannis Montreuil and Hoang Duy Dang and Maxime Meyer and Lai Xing Ng and Axel Carlier and Wei Tsang Ooi},
booktitle={The 29th International Conference on Artificial Intelligence and Statistics},
year={2026},
url={https://openreview.net/forum?id=1lix8ppUJ7}
}

@article{montreuil2026learningdefernonstationarytime,
  title={Learning to Defer in Non-Stationary Time Series via Switching State-Space Models},
  author={Yannis Montreuil and Letian Yu and Axel Carlier and Lai Xing Ng and Wei Tsang Ooi},
  journal={arXiv preprint arXiv:2601.22538},
  year={2026}
}

@article{Statistical,
author = {Zhang, Tong},
year = {2002},
month = {12},
pages = {},
title = {Statistical Behavior and Consistency of Classification Methods based on Convex Risk Minimization},
volume = {32},
journal = {Annals of Statistics},
doi = {10.1214/aos/1079120130}
}

@inproceedings{Mozannar2023WhoSP,
  title={Who Should Predict? Exact Algorithms For Learning to Defer to Humans},
  author={Hussein Mozannar and Hunter Lang and Dennis Wei and Prasanna Sattigeri and Subhro Das and David A. Sontag},
  booktitle={International Conference on Artificial Intelligence and Statistics},
  year={2023},
  url={https://api.semanticscholar.org/CorpusID:255941521}
}

@inproceedings{charusaie2022sample,
  title={Sample efficient learning of predictors that complement humans},
  author={Charusaie, Mohammad-Amin and Mozannar, Hussein and Sontag, David and Samadi, Samira},
  booktitle={International Conference on Machine Learning},
  pages={2972--3005},
  year={2022},
  organization={PMLR}
}

@article{tewari07a,
  author  = {Ambuj Tewari and Peter L. Bartlett},
  title   = {On the Consistency of Multiclass Classification Methods},
  journal = {Journal of Machine Learning Research},
  year    = {2007},
  volume  = {8},
  number  = {36},
  pages   = {1007--1025},
  url     = {http://jmlr.org/papers/v8/tewari07a.html}
}

@misc{communities_and_crime_183,
  author       = {Redmond, Michael},
  title        = {{Communities and Crime}},
  year         = {2002},
  howpublished = {UCI Machine Learning Repository},
  note         = {{DOI}: https://doi.org/10.24432/C53W3X}
}

@article{dermamnist1,
    title = {The HAM10000 dataset, a large collection of multi-source dermatoscopic images of common pigmented skin lesions},
    author = {Tschandl, Philipp and Rosendahl, Cliff and Kittler, Harald},
    journal = {Scientific data},
    pages = {180161},
    year = {2018},
    publisher = {Nature Publishing Group}
}

@article{dermamnist2,
    title = {Skin lesion analysis toward melanoma detection 2018: A challenge hosted by the international skin imaging collaboration (isic)},
    author = {Codella, Noel and Rotemberg, Veronica and Tschandl, Philipp and Celebi, M Emre and Dusza, Stephen and Gutman, David and Helba, Brian and Kalloo, Aadi and Liopyris, Konstantinos and Marchetti, Michael and others},
    journal = {arXiv preprint arXiv:1902.03368},
    year = {2019}
}

@article{medmnistv2,
    title={MedMNIST v2-A large-scale lightweight benchmark for 2D and 3D biomedical image classification},
    author={Yang, Jiancheng and Shi, Rui and Wei, Donglai and Liu, Zequan and Zhao, Lin and Ke, Bilian and Pfister, Hanspeter and Ni, Bingbing},
    journal={Scientific Data},
    volume={10},
    number={1},
    pages={41},
    year={2023},
    publisher={Nature Publishing Group UK London}
}

@inproceedings{medmnistv1,
    title={MedMNIST Classification Decathlon: A Lightweight AutoML Benchmark for Medical Image Analysis},
    author={Yang, Jiancheng and Shi, Rui and Ni, Bingbing},
    booktitle={IEEE 18th International Symposium on Biomedical Imaging (ISBI)},
    pages={191--195},
    year={2021}
}

@article{montreuil2026learningtodeferexpertconditionedadvice,
  title={Learning-to-Defer with Expert-Conditioned Advice},
  author={Montreuil, Yannis and Montreuil, Le{\"\i}na and Carlier, Axel and Ng, Lai Xing and Ooi, Wei Tsang},
  journal={arXiv preprint arXiv:2603.14324},
  year={2026}
}

@article{montreuil2026beyond,
  title={Beyond Augmented-Action Surrogates for Multi-Expert Learning-to-Defer},
  author={Montreuil, Yannis and Carlier, Axel and Ng, Lai Xing and Ooi, Wei Tsang},
  journal={arXiv preprint arXiv:2604.09414},
  year={2026}
}

@article{mao2024h,
  title={$ H $-Consistency Bounds: Characterization and Extensions},
  author={Mao, Anqi and Mohri, Mehryar and Zhong, Yutao},
  journal={Advances in Neural Information Processing Systems},
  volume={36},
  year={2023}
}

@inproceedings{theoretically,
  title = 	 {Theoretically Grounded Loss Functions and Algorithms for Score-Based Multi-Class Abstention},
  author =       {Mao, Anqi and Mohri, Mehryar and Zhong, Yutao},
  booktitle = 	 {Proceedings of The 27th International Conference on Artificial Intelligence and Statistics},
  pages = 	 {4753--4761},
  year = 	 {2024},
  editor = 	 {Dasgupta, Sanjoy and Mandt, Stephan and Li, Yingzhen},
  volume = 	 {238},
  series = 	 {Proceedings of Machine Learning Research},
  month = 	 {02--04 May},
  publisher =    {PMLR},
  url = 	 {https://proceedings.mlr.press/v238/mao24a.html}
}

@inproceedings{mao2024principledapproacheslearningdefer,
  title={Principled Approaches for Learning to Defer with Multiple Experts},
  author={Mao, Anqi and Mohri, Mehryar and Zhong, Yutao},
  booktitle={International Symposium on Artificial Intelligence and Mathematics},
  year={2024}
}

@inproceedings{awasthi2021calibrationconsistencyadversarialsurrogate,
  title={Calibration and Consistency of Adversarial Surrogate Losses},
  author={Awasthi, Pranjal and Frank, Natalie and Mao, Anqi and Mohri, Mehryar and Zhong, Yutao},
  booktitle={Advances in Neural Information Processing Systems},
  volume={34},
  year={2021}
}

@article{mao2024realizablehconsistentbayesconsistentloss,
  title={Realizable $ H $-Consistent and Bayes-Consistent Loss Functions for Learning to Defer},
  author={Mao, Anqi and Mohri, Mehryar and Zhong, Yutao},
  journal={Advances in neural information processing systems},
  volume={37},
  pages={73638--73671},
  year={2024}
}

@InProceedings{Grounded,
  title = 	 {Theoretically Grounded Loss Functions and Algorithms for Adversarial Robustness},
  author =       {Awasthi, Pranjal and Mao, Anqi and Mohri, Mehryar and Zhong, Yutao},
  booktitle = 	 {Proceedings of The 26th International Conference on Artificial Intelligence and Statistics},
  pages = 	 {10077--10094},
  year = 	 {2023},
  editor = 	 {Ruiz, Francisco and Dy, Jennifer and van de Meent, Jan-Willem},
  volume = 	 {206},
  series = 	 {Proceedings of Machine Learning Research},
  month = 	 {25--27 Apr},
  publisher =    {PMLR},
  url = 	 {https://proceedings.mlr.press/v206/awasthi23c.html}
}

@inproceedings{mao2024regressionmultiexpertdeferral,
author = {Mao, Anqi and Mohri, Mehryar and Zhong, Yutao},
title = {Regression with multi-expert deferral},
year = {2024},
publisher = {JMLR.org},
booktitle = {Proceedings of the 41st International Conference on Machine Learning},
articleno = {1413},
numpages = {22},
location = {Vienna, Austria},
series = {ICML'24}
}

@inproceedings{Mao_Mohri_Zhong_2023,
  title={Predictor-rejector multi-class abstention: Theoretical analysis and algorithms},
  author={Mao, Anqi and Mohri, Mehryar and Zhong, Yutao},
  booktitle={International Conference on Algorithmic Learning Theory},
  pages={822--867},
  year={2024},
  organization={PMLR}
}

@inproceedings{mao2023crossentropylossfunctionstheoretical,
  title={Cross-entropy loss functions: Theoretical analysis and applications},
  author={Mao, Anqi and Mohri, Mehryar and Zhong, Yutao},
  booktitle={International conference on Machine learning},
  pages={23803--23828},
  year={2023},
  organization={PMLR}
}

@inproceedings{
cortes2024cardinalityaware,
title={Cardinality-Aware Set Prediction and Top-\$k\$ Classification},
author={Corinna Cortes and Anqi Mao and Christopher Mohri and Mehryar Mohri and Yutao Zhong},
booktitle={The Thirty-eighth Annual Conference on Neural Information Processing Systems},
year={2024},
url={https://openreview.net/forum?id=WAT3qu737X}
}

@inproceedings{Awasthi_Mao_Mohri_Zhong_2022_multi,
author = {Awasthi, Pranjal and Mao, Anqi and Mohri, Mehryar and Zhong, Yutao},
title = {Multi-class H-consistency bounds},
year = {2022},
isbn = {9781713871088},
publisher = {Curran Associates Inc.},
address = {Red Hook, NY, USA},
booktitle = {Proceedings of the 36th International Conference on Neural Information Processing Systems},
articleno = {57},
numpages = {14},
location = {New Orleans, LA, USA},
series = {NIPS '22}
}

@inproceedings{mao2023twostage,
title={Two-Stage Learning to Defer with Multiple Experts},
author={Anqi Mao and Christopher Mohri and Mehryar Mohri and Yutao Zhong},
booktitle={Thirty-seventh Conference on Neural Information Processing Systems},
year={2023},
url={https://openreview.net/forum?id=GIlsH0T4b2}
}

@inproceedings{mohri2026beyond,
  title={Beyond Tsybakov: Model Margin Noise and {H}-Consistency Bounds},
  author={Mohri, Mehryar and Zhong, Yutao},
  booktitle={International Symposium on Artificial Intelligence and Mathematics},
  year={2026}
}

@inproceedings{mohri2026mind,
  title={Mind the Gap: Structure-Aware Consistency in Preference Learning},
  author={Mohri, Mehryar and Zhong, Yutao},
  booktitle={Forty-Third International Conference on Machine Learning},
  year={2026}
}

@inproceedings{mohri2026linear,
  title={Linear-Core Surrogates: Smooth Loss Functions with Linear Rates for Classification and Structured Prediction},
  author={Mohri, Mehryar and Zhong, Yutao},
  booktitle={Forty-Third International Conference on Machine Learning},
  year={2026}
}

@inproceedings{cortes2026optimized,
  title={Optimized Deferral for Imbalanced Settings},
  author={Cortes, Corinna and Mao, Anqi and Mohri, Mehryar and Zhong, Yutao},
  booktitle={Forty-Third International Conference on Machine Learning},
  year={2026}
}

@inproceedings{mao2025mastering,
  title={Mastering Multiple-Expert Routing: Realizable {$H$}-Consistency and Strong Guarantees for Learning to Defer},
  author={Mao, Anqi and Mohri, Mehryar and Zhong, Yutao},
  booktitle={Forty-second International Conference on Machine Learning},
  year={2025}
}

@article{desalvo2025budgeted,
  title={Budgeted Multiple-Expert Deferral},
  author={DeSalvo, Giulia and Mohri, Clara and Mohri, Mehryar and Zhong, Yutao},
  journal={arXiv preprint arXiv:2510.26706},
  year={2025}
}

@inproceedings{cortes2026theoretical,
  title={A Theoretical Framework for Modular Learning of Robust Generative Models},
  author={Cortes, Corinna and Mohri, Mehryar and Zhong, Yutao},
  booktitle={Forty-Third International Conference on Machine Learning},
  year={2026}
}

@article{mohri2026generalized,
  title={Generalized Distributional Alignment Games for Unbiased Answer-Level Fine-Tuning},
  author={Mohri, Mehryar and Schneider, Jon and Zhong, Yutao},
  journal={arXiv preprint arXiv:2605.02435},
  year={2026}
}

@article{mohri2026principled,
  title={Principled Algorithms for Optimizing Generalized Metrics in Multi-Label Learning},
  author={Mohri, Mehryar and Zhong, Yutao},
  journal={arXiv preprint arXiv:2605.28767},
  year={2026}
}

@inproceedings{mohri2024learningreject,
  title={Learning to Reject with a Fixed Predictor: Application to Decontextualization},
  author={Mohri, Christopher and Andor, Daniel and Choi, Eunsol and Collins, Michael and Mao, Anqi and Zhong, Yutao},
  booktitle={The Twelfth International Conference on Learning Representations},
  year={2024}
}

@inproceedings{mao2023rankingabstention,
  title={Ranking with Abstention},
  author={Mao, Anqi and Mohri, Mehryar and Zhong, Yutao},
  booktitle={ICML 2023 Workshop on The Many Facets of Preference-Based Learning},
  year={2023}
}

@inproceedings{mao2023structuredprediction,
  title={Structured Prediction with Stronger Consistency Guarantees},
  author={Mao, Anqi and Mohri, Mehryar and Zhong, Yutao},
  booktitle={Advances in Neural Information Processing Systems},
  year={2023}
}

@inproceedings{mao2023pairwisemisranking,
  title={{H}-Consistency Bounds for Pairwise Misranking Loss Surrogates},
  author={Mao, Anqi and Mohri, Mehryar and Zhong, Yutao},
  booktitle={Proceedings of the 40th International Conference on Machine Learning},
  pages={23743--23802},
  year={2023}
}

@inproceedings{mao2025principledbinary,
  title={Principled Algorithms for Optimizing Generalized Metrics in Binary Classification},
  author={Mao, Anqi and Mohri, Mehryar and Zhong, Yutao},
  booktitle={Forty-second International Conference on Machine Learning},
  year={2025}
}

@inproceedings{cortes2025improvedbalanced,
  title={Improved Balanced Classification with Theoretically Grounded Loss Functions},
  author={Cortes, Corinna and Mohri, Mehryar and Zhong, Yutao},
  booktitle={Advances in Neural Information Processing Systems},
  year={2025}
}

@phdthesis{zhong2025thesis,
  title={Fundamental Novel Consistency Theory: {H}-Consistency Bounds},
  author={Zhong, Yutao},
  school={New York University},
  year={2025}
}

@inproceedings{cortes2025balancingscales,
  title={Balancing the Scales: A Theoretical and Algorithmic Framework for Learning from Imbalanced Data},
  author={Cortes, Corinna and Mao, Anqi and Mohri, Mehryar and Zhong, Yutao},
  booktitle={Forty-second International Conference on Machine Learning},
  year={2025}
}

@inproceedings{mao2025enhanced,
  title={Enhanced \$H\$-Consistency Bounds},
  author={Anqi Mao and Mehryar Mohri and Yutao Zhong},
  booktitle={36th International Conference on Algorithmic Learning Theory},
  year={2025},
  url={https://openreview.net/forum?id=qgnVGFJMJo}
}

@inproceedings{mao2024universalgrowth,
  title={A Universal Growth Rate for Learning with Smooth Surrogate Losses},
  author={Mao, Anqi and Mohri, Mehryar and Zhong, Yutao},
  booktitle={Advances in Neural Information Processing Systems},
  year={2024}
}

@inproceedings{mao2024hconsistencyregression,
  title={{H}-Consistency Guarantees for Regression},
  author={Mao, Anqi and Mohri, Mehryar and Zhong, Yutao},
  booktitle={Forty-first International Conference on Machine Learning},
  year={2024}
}

@inproceedings{mao2024multilabel,
  title={Multi-Label Learning with Stronger Consistency Guarantees},
  author={Mao, Anqi and Mohri, Mehryar and Zhong, Yutao},
  booktitle={Advances in Neural Information Processing Systems},
  year={2024}
}

@phdthesis{mao2025thesis,
  title={Theory and Algorithms for Learning with Multi-Class Abstention and Multi-Expert Deferral},
  author={Mao, Anqi},
  school={New York University},
  year={2025}
}

%%%%%%%%%%%%%%%%%%%%%%%%%%%%%%%%%%%%%%%%%%%%%%%%%%%%%%%%%%%%
\section*{Checklist}

 \begin{enumerate}

 \item For all models and algorithms presented, check if you include:
 \begin{enumerate}
   \item A clear description of the mathematical setting, assumptions, algorithm, and/or model. \textcolor{aistatsblue}{(Yes) see Appendix \ref{assumptions}}
   \item An analysis of the properties and complexity (time, space, sample size) of any algorithm. \textcolor{aistatsblue}{(Yes) see Appendix \ref{complexity}}
   \item (Optional) Anonymized source code, with specification of all dependencies, including external libraries. \textcolor{aistatsblue}{(Yes)}
 \end{enumerate}

 \item For any theoretical claim, check if you include:
 \begin{enumerate}
   \item Statements of the full set of assumptions of all theoretical results. \textcolor{aistatsblue}{(Yes) see Appendix \ref{assumptions}}
   \item Complete proofs of all theoretical results. \textcolor{aistatsblue}{(Yes) see Appendix}
   \item Clear explanations of any assumptions. \textcolor{aistatsblue}{(Yes) see Appendix \ref{assumptions}}
 \end{enumerate}

 \item For all figures and tables that present empirical results, check if you include:
 \begin{enumerate}
   \item The code, data, and instructions needed to reproduce the main experimental results (either in the supplemental material or as a URL). \textcolor{aistatsblue}{(Yes)}
   \item All the training details (e.g., data splits, hyperparameters, how they were chosen). \textcolor{aistatsblue}{(Yes)}
         \item A clear definition of the specific measure or statistics and error bars (e.g., with respect to the random seed after running experiments multiple times). \textcolor{aistatsblue}{(Yes)}
         \item A description of the computing infrastructure used. (e.g., type of GPUs, internal cluster, or cloud provider). \textcolor{aistatsblue}{(Yes)}
 \end{enumerate}

 \item If you are using existing assets (e.g., code, data, models) or curating/releasing new assets, check if you include:
 \begin{enumerate}
   \item Citations of the creator If your work uses existing assets. \textcolor{aistatsblue}{(Yes)}
   \item The license information of the assets, if applicable. \textcolor{aistatsblue}{(Not applicable)}
   \item New assets either in the supplemental material or as a URL, if applicable. \textcolor{aistatsblue}{(Not applicable)}
   \item Information about consent from data providers/curators. \textcolor{aistatsblue}{(Not applicable)}
   \item Discussion of sensible content if applicable, e.g., personally identifiable information or offensive content. \textcolor{aistatsblue}{(Not applicable)}
 \end{enumerate}

 \item If you used crowdsourcing or conducted research with human subjects, check if you include:
 \begin{enumerate}
   \item The full text of instructions given to participants and screenshots. \textcolor{aistatsblue}{(Not applicable)}
   \item Descriptions of potential participant risks, with links to Institutional Review Board (IRB) approvals if applicable. \textcolor{aistatsblue}{(Not applicable)}
   \item The estimated hourly wage paid to participants and the total amount spent on participant compensation. \textcolor{aistatsblue}{(Not applicable)}
 \end{enumerate}
 \end{enumerate}

\newpage
\clearpage
\onecolumn
\appendix 
\setcounter{table}{0}
\renewcommand{\thetable}{A\arabic{table}}
\renewcommand{\theHtable}{A\arabic{table}}
\newpage

\section{Appendix}

\subsection{Important Definitions, Lemmas, and Theorems} \label{assumptions}

\begin{definition}[Symmetric Hypothesis Class]
\label{def:symmetric}
Let $\mc{A}$ denote the set of possible actions (predictions and deferrals), and let $Q$ be a class of score-valued hypotheses $q:\mc{X}\to\mathbb{R}^{|\mc{A}|}$. 
We say that $Q$ is \emph{symmetric} if it is closed under permutations of the coordinates indexed by $\mc{A}$, i.e., for any $q \in Q$ and any permutation $\Pi:\mc{A}\to\mc{A}$, the permuted hypothesis $q^\Pi:\mc{X}\to\mathbb{R}^{|\mc{A}|}$ defined by
\[
q^\Pi(x)_j = q(x)_{\Pi^{-1}(j)}, \qquad \forall x \in \mc{X}, \ \forall j \in \mc{A},
\]
also belongs to $Q$. 
\end{definition}

\begin{restatable}[Locally \(\rho\)-consistent \citep{Grounded}]{definition}{rhoconsistency}\label{rho_consistent}
A hypothesis set \( \mathcal{Q} \) is \textit{locally \(\rho\)-consistent} if, for any \( x \in \mathcal{X} \), there exists a hypothesis \( q \in \mathcal{Q} \) such that:
\[
\inf_{x' \in B_p(x, \gamma)} |q(x')_i - q(x')_j| \geq \rho,
\]
where \( \rho > 0 \), \( i \neq j \in \mathcal{A} \), and \( x' \in B_p(x, \gamma) \). Moreover, the ordering of the values \( \{q(x')_j\} \) is preserved with respect to \( \{q(x)_j\} \) for all \( x' \in B_p(x, \gamma) \).
\end{restatable}

\begin{restatable}[$\mathcal{Q}$-consistency bounds from \citet{montreuil2025adversarial}]{lemma}{qconsistency}\label{lemma:qconsistency} 
Assume \( \mathcal{Q} \) is symmetric and locally \( \rho \)-consistent. Then, for the set \( \mathcal{A} \), any hypothesis \( q \in \mathcal{Q} \), and any distribution \( \mathcal{P} \) with probabilities \( p = (p_1, \ldots, p_{|\mc{A}|}) \in \Delta^{|\mathcal{A}|} \), the following inequality holds:
\begin{equation*}
\begin{aligned}
     \sum_{j \in \mathcal{A}} p_j &\sup_{x_j'\in B_p(x,\gamma)} \mathbf{1}\{\hat{q}(x_j')\not=j\} - \inf_{q \in \mathcal{Q}} \sum_{j \in \mathcal{A}} p_j \sup_{x_j'\in B_p(x,\gamma)}\mathbf{1}\{\hat{q}(x_j')\not=j\} \leq \\
    & \Psi^u(1) \Big( \sum_{j \in \mathcal{A}} p_j \sup_{x_j'\in B_p(x,\gamma)}\Phi^{\rho,u}_{\text{cls}}(q(x_j'), j) - \inf_{q \in \mathcal{Q}} \sum_{j \in \mathcal{A}} p_j \sup_{x_j'\in B_p(x,\gamma)}\Phi^{\rho,u}_{\text{cls}}(q(x_j'), j) \Big).
\end{aligned}
\end{equation*}
with $\Psi^u(1) =
\begin{cases}
\log(2), & u=1, \\[0.3em]
\tfrac{1}{1-u}\big(2^{1-u}-1\big), & u \neq 1.
\end{cases}$
\end{restatable}

\subsection{Proof of Lemma \ref{lemma:surrogate_class}} \label{proof:lemma:surrogate_class}
\surrogatemulti* 

\begin{proof}
Fix a target class $j\in\mathcal A$. Define
\[
\Phi^{\rho,u}_{\text{cls}}(h(x),j)
=\Psi^u\left(\sum_{\substack{j'\in\mathcal A\\ j'\neq j}}
\Psi_\rho\big(h(x)_{j'}-h(x)_j\big)\right),
\qquad
\widetilde{\Phi}^{\rho,u}_{\text{cls}}(h,x,j)
=\sup_{x'_j\in B_p(x,\gamma)}
\Psi^u\left(\sum_{\substack{j'\in\mathcal A\\ j'\neq j}}
\Psi_\rho\big(h_{j'}(x'_j)-h_j(x'_j)\big)\right).
\]
We take the exponential link $\Psi_{\mathrm e}(v)=e^{-v}$ and the $\rho$-softening
\[
\Psi_\rho(v)=\Psi_{\mathrm e}\Big(\frac{v}{\rho}\Big)=\exp\Big(-\frac{v}{\rho}\Big),\qquad \rho>0.
\]
For $u>0$ define
\[
\Psi^{u}(v)=
\begin{cases}
\log(1+v), & u=1,\\[2pt]
\dfrac{(1+v)^{1-u}-1}{1-u}, & u\neq 1,
\end{cases}
\qquad v\ge 0.
\]
Then $\Psi^{u}$ is nondecreasing and $1$-Lipschitz on $\mathbb R_+$ since
$\big|\frac{\partial}{\partial v}\Psi^u(v)\big|=\frac{1}{(1+v)^u}\le 1$ for $v\ge 0$ and $u>0$.
Moreover, $\Psi_\rho$ is $\frac{1}{\rho}$-Lipschitz on $\mathbb R$.

For $j'\neq j$ define the pairwise margin
\[
\Delta_h(x,j,j')\ =\ h(x)_j-h(x)_{j'}.
\]
Collect these $|\mc{A}|-1$ margins into
\[
\overline{\Delta}_h(x,j)
=\big(\Delta_h(x,j,1),\ldots,\Delta_h(x,j,j-1),\Delta_h(x,j,j+1),\ldots,\Delta_h(x,j,|\mc{A}|)\big)\in\mathbb R^{|\mc{A}|-1}.
\]
Note that $h(x)_{j'}-h(x)_j=-\Delta_h(x,j,j')$, hence
$\Psi_\rho\big(h_{j'}(\cdot)-h_j(\cdot)\big)=\Psi_\rho\big(-\Delta_h(\cdot,j,j')\big)$.

Using monotonicity and $1$-Lipschitzness of $\Psi^u$ on $\mathbb R_+$,
\begin{align}
\widetilde{\Phi}^{\rho,u}_{\text{cls}}(h,x,j)
&=\sup_{x'_j\in B_p(x,\gamma)}\ \Psi^u\left(\sum_{j'\neq j}\Psi_\rho\big(-\Delta_h(x'_j,j,j')\big)\right)\nonumber\\
&\le \Psi^u\left(\sum_{j'\neq j}\Psi_\rho\big(-\Delta_h(x,j,j')\big)\right)
+\sup_{x'_j\in B_p(x,\gamma)}
\left|\sum_{j'\neq j}\Big(\Psi_\rho\big(-\Delta_h(x'_j,j,j')\big)-\Psi_\rho\big(-\Delta_h(x,j,j')\big)\Big)\right|\nonumber\\
&=\Phi^{\rho,u}_{\text{cls}}(h(x),j)\ +\ \sup_{x'_j\in B_p(x,\gamma)}
\sum_{j'\neq j}\left|\Psi_\rho\big(-\Delta_h(x'_j,j,j')\big)-\Psi_\rho\big(-\Delta_h(x,j,j')\big)\right|.
\label{eq:step1-h}
\end{align}

Since $\Psi_\rho$ is $\frac{1}{\rho}$-Lipschitz and by Cauchy--Schwarz,
\begin{align}
\sum_{j'\neq j}\left|\Psi_\rho\big(-\Delta_h(x'_j,j,j')\big)-\Psi_\rho\big(-\Delta_h(x,j,j')\big)\right|
&\le \frac{1}{\rho}\sum_{j'\neq j}\big|\Delta_h(x'_j,j,j')-\Delta_h(x,j,j')\big|\nonumber\\
&\le \frac{\sqrt{|\mc{A}|-1}}{\rho}\,\big\|\overline{\Delta}_h(x'_j,j)-\overline{\Delta}_h(x,j)\big\|_2.
\label{eq:step2-h}
\end{align}
Plugging \eqref{eq:step2-h} into \eqref{eq:step1-h} gives, for all $j$,
\begin{equation}
\label{eq:key-bound-h}
\widetilde{\Phi}^{\rho,u}_{\text{cls}}(h,x,j)
\ \le\ \Phi^{\rho,u}_{\text{cls}}(h(x),j)\ +\ \kappa\ \sup_{x'_j\in B_p(x,\gamma)}
\big\|\overline{\Delta}_h(x'_j,j)-\overline{\Delta}_h(x,j)\big\|_2,
\qquad
\kappa\geq\frac{\sqrt{|\mc{A}|-1}}{\rho}.
\end{equation}

Because $\Psi_\rho(v)=\exp(-v/\rho)=\Psi_{\mathrm e}(v/\rho)$ and $\Psi^u$ is nondecreasing,
\[
\Phi^{\rho,u}_{\text{cls}}(h(x),j)
=\Psi^u\left(\sum_{j'\neq j}\Psi_{\mathrm e}\left(\frac{h(x)_{j'}-h(x)_j}{\rho}\right)\right)
=\Psi^u\left(\sum_{j'\neq j}\Psi_{\mathrm e}\left(\frac{h(x)_{j'}}{\rho}-\frac{h(x)_j}{\rho}\right)\right)
=\ \Phi^{u}_{\text{cls}}\left(\frac{h}{\rho},x,j\right),
\]
where $\frac{h}{\rho}$ denotes the score map $x\mapsto h(x)/\rho$ (componentwise).
Hence \eqref{eq:key-bound-h} becomes
\begin{equation}
\label{eq:final-upper-h}
\widetilde{\Phi}^{\rho,u}_{\text{cls}}(h,x,j)
\ \le\ \Phi^{u}_{\text{cls}}\left(\frac{h}{\rho},x,j\right)\ +\ \kappa\ \sup_{x'_j\in B_p(x,\gamma)}
\big\|\overline{\Delta}_h(x'_j,j)-\overline{\Delta}_h(x,j)\big\|_2,
\end{equation}

Define the smooth upper-bounding adversarial surrogate
\[
\widetilde{\Phi}^{\mathrm{smth},u}_{\text{cls}}(h,x,j)
=\ \Phi^{u}_{\text{cls}}\left(\frac{h}{\rho},x,j\right)\ +\ \kappa\ \sup_{x'_j\in B_p(x,\gamma)}
\big\|\overline{\Delta}_h(x'_j,j)-\overline{\Delta}_h(x,j)\big\|_2,
\]
Then \eqref{eq:final-upper-h} states, pointwise in $(x,j)$,
\[
\widetilde{\Phi}^{\rho,u}_{\text{cls}}(h,x,j)\ \le\ \widetilde{\Phi}^{\mathrm{smth},u}_{\text{cls}}(h,x,j).
\]
\end{proof}

\subsection{Proof of Lemma \ref{lem:tlcrobustness}} \label{proof:lem:tlcrobustness}
\tlcrobustness*

\begin{proof}
    We first show that we can rewrite the usual score-based loss \citep{mozannar2021consistent, mao2024principledapproacheslearningdefer} introduced in Equation \ref{class} with a cost-sensitive formulation depending on a shifted cost $\mu_j$. Let the following change of variable:
    \begin{equation}\label{eq:1}
        \mu_j(j,m_j,y) = \begin{cases}
            \alpha_j\mathbf{1}\{j\not=y\} + \beta_j & \text{if } j\leq K \\
            \alpha_j\mathbf{1}\{m_{j-K}\not=y\} + \beta_j & \text{if } j> K, \\
        \end{cases}
    \end{equation}
leading to
\begin{equation}
    \ell_{\text{def}}^c(\hat{h}(x),y,\mathbf{m}) = \sum_{j=1}^{K+J}\mu_j(j,m_j,y)\mathbf{1}\{\hat{h}(x)=j\}.
\end{equation} 

It is immediate that the formulation reduces to the standard score-based loss when setting $\alpha_j=1$ and $\beta_j=0$ for $j \leq K$. For instance, if $\mc{Y}=\{1,2\}$ and the prediction is $\hat{h}(x)=1$, then by Equation~\ref{eq:1} we obtain $\ell_{\mathrm{def}}^c(\hat{h}(x),y,\mathbf{m})=\mathbf{1}\{1\neq y\}$, which coincides with the usual score-based loss in Equation~\ref{class}.

Now, we upper-bound this loss with the outcome-wise worst-case scenario under the newly cost-sensitive formulation:
\begin{equation}
    \begin{aligned}
        \ell_{\text{def}}^c(\hat{h}(x),y,\mathbf{m}) & \leq \sup_{x' \in B_p(x,\gamma)} \sum_{j=1}^{K+J}\mu_j(j,m_j,y)\mathbf{1}\{\hat{h}(x')=j\} \\
        & \leq  \sum_{j=1}^{K+J}\mu_j(j,m_j,y)\sup_{x'_j \in B_p(x,\gamma)}\mathbf{1}\{\hat{h}(x'_j)=j\} \\
        & = \widetilde{\ell}_{\text{def}}^c(h,x,y,\mathbf{m})
    \end{aligned}
\end{equation}

\end{proof}

\subsection{Proof of Theorem \ref{theo:consistency}}\label{proof:theo:consistency}
\consistency*

\begin{proof}Let \(\mathcal{A}=\{1,\dots,K{+}J\}\). For \(x\in\mathcal{X}\) and radius \(\gamma>0\), write
\(B_p(x,\gamma)=\{x'\in\mathcal{X}:\|x'-x\|_p\le\gamma\}\).
Given a score vector \(h(x)\in\mathbb{R}^{K+J}\), let the induced decision be
\(\hat h(x)\in\mathcal{A}\) (e.g., argmax with fixed tie-breaking). 
We begin by recalling the outcome-wise adversarial deferral loss
\begin{equation}
    \widetilde{\ell}_{\mathrm{def}}^c(h,x,y,\mathbf{m}) 
    = \sum_{j=1}^{K+J} \mu_j(j,m_j,y)\,
      \sup_{x'_j \in B_p(x,\gamma)} \mathbf{1}\{\hat{h}(x'_j)=j\},
\end{equation}
together with its surrogate counterpart introduced in Definition~\ref{lem:scrobustness}:
\begin{equation}
    \widetilde{\Phi}_{\mathrm{def}}^{c,u}(h,x,y,\mathbf{m}) 
    = \sum_{j=1}^{K+J} \sum_{i\neq j} \mu_i(i,m_i,y)\,
      \sup_{x'_j \in B_p(x,\gamma)} \Phi_{\mathrm{cls}}^{\rho,u}\big(h(x'_j),j\big).
\end{equation}
\paragraph{True Loss Calibration Gap.} 
Let define the conditional risk associated with the adversarial true loss:
\begin{equation}
    \begin{aligned}
        \mc{C}_{\widetilde{\ell}_{\mathrm{def}}^c}(h,x) 
    & = \mathbb{E}_{Y\mid X=x}\,\mathbb{E}_{M\mid X=x,Y}\big[\widetilde{\ell}_{\mathrm{def}}^c(h,x,Y,M)\big] \\
    & = \sum_{j=1}^{K+J} \mathbb{E}_{Y\mid X=x}\,\mathbb{E}_{M_j\mid X=x,Y}\Big[\mu_j(j,M_j,Y)\Big]\sup_{x'_j \in B_p(x,\gamma)} \mathbf{1}\{\hat{h}(x'_j)=j\} \\
    & = \sum_{j=1}^{K+J} \overline{\mu}_j(x)\sup_{x'_j \in B_p(x,\gamma)} \mathbf{1}\{\hat{h}(x'_j)=j\}
    \end{aligned}
\end{equation}
with $\overline{\mu}_j(x) = \mathbb{E}_{Y\mid X=x}\,\mathbb{E}_{M_j\mid X=x,Y}\Big[\mu_j(j,M_j,Y)\Big]$. Next, we assume there exist a function $h \in \mc{H}$ that is local-$\rho$-consistent (see Lemma \ref{rho_consistent}). For any \(h\) and \(x\), define the reachability set
\begin{equation}
    \mathcal{H}_\gamma(h,x)\;=\;\Big\{j\in\mathcal{A}:\ \exists\,x'_j\in B_p(x,\gamma)\ \text{s.t.}\ \hat h(x'_j)=j\Big\}.
\end{equation}
Then, for each \(j\in\mathcal{A}\),
\begin{equation}
\sup_{x'_j\in B_p(x,\gamma)}\mathbf{1}\{\hat h(x'_j)=j\}
=\mathbf{1}\{\,j\in\mathcal{H}_\gamma(h,x)\,\}.
\end{equation}
By definition, the supremum over \(x'_j\in B_p(x,\gamma)\) of the indicator equals \(1\) iff
there exists at least one \(x'_j\) in the ball with \(\hat h(x'_j)=j\), i.e., iff \(j\in\mathcal{H}_\gamma(h,x)\);
otherwise it is \(0\). Consequently,
\begin{equation}
\mathcal{C}_{\widetilde{\ell}_{\mathrm{def}}^c}(h,x)
=\sum_{j=1}^{K+J}\overline{\mu}_j(x)\,\mathbf{1}\{\,j\in\mathcal{H}_\gamma(h,x)\,\}.
\end{equation}
For any \(h\), \(\mathcal{H}_\gamma(h,x)\neq\varnothing\) because \(x\in B_p(x,\gamma)\) and \(\hat h(x)\in\mathcal{A}\),
hence at least one label is realized in the ball.
\begin{equation}
\mathcal{C}_{\widetilde{\ell}_{\mathrm{def}}^c}(h,x)
=\sum_{j\in\mathcal{A}}\overline{\mu}_j(x)\,\mathbf{1}\{j\in\mathcal{H}_\gamma(h,x)\}
\ \ge\ \min_{j\in\mathcal{A}}\overline{\mu}_j(x)
\mathbf{1}\{\mathcal{H}_\gamma(h,x)\neq\varnothing\}
=\min_{j\in\mathcal{A}}\overline{\mu}_j(x).
\end{equation}
Let \(j^\star\in\arg\min_{j\in\mathcal{A}}\overline{\mu}_j(x)\).
By local-$\rho$-consistency, there exists \(h^{j^\star}\) with \(\hat h^{j^\star}(x')=j^\star\) for all \(x'\in B_p(x,\gamma)\).
Therefore \(\mathcal{H}_\gamma(h^{j^\star},x)=\{j^\star\}\).
\begin{equation}
\mathcal{C}_{\widetilde{\ell}_{\mathrm{def}}^c}(h^{j^\star},x)
=\sum_{j\in\mathcal{A}}\overline{\mu}_j(x)\,\mathbf{1}\{j=j^\star\}
=\overline{\mu}_{j^\star}(x)
=\min_{j\in\mathcal{A}}\overline{\mu}_j(x).
\end{equation}
Combining the lower bound with achievability yields
\begin{equation}
\inf_{h \in \mc{H}}\ \mathcal{C}_{\widetilde{\ell}_{\mathrm{def}}^c}(h,x)
\;=\mathcal{C}^B_{\widetilde{\ell}_{\mathrm{def}}^c}(\mc{H},x) \;= \;\min_{j\in\mathcal{A}}\overline{\mu}_j(x).
\end{equation}
Next, let's define the calibration gap:
\begin{equation}
    \begin{aligned}
        \Delta\mathcal{C}_{\widetilde{\ell}_{\mathrm{def}}^c}(h,x) & = \mathcal{C}_{\widetilde{\ell}_{\mathrm{def}}^c}(h,x) - \mathcal{C}^B_{\widetilde{\ell}_{\mathrm{def}}^c}(\mc{H},x) \\
        & = \mathcal{C}_{\widetilde{\ell}_{\mathrm{def}}^c}(h,x) - \min_{j\in\mc{A}}\overline{\mu}_j(x) \\
        & = \sum_{j=1}^{K+J}\Big( \overline{\mu}_j(x)\sup_{x'_j \in B_p(x,\gamma)} \mathbf{1}\{\hat{h}(x'_j)=j\}\Big) - \min_{j\in\mc{A}}\overline{\mu}_j(x)
    \end{aligned}
\end{equation}

\paragraph{Surrogate Calibration Gap.} 
Now, for the surrogate, we have:
\begin{equation}
    \begin{aligned}
        \mathcal{C}_{\widetilde{\Phi}_{\mathrm{def}}^{c,u}}(h,x) = \sum_{j=1}^{K+J} \sum_{i\not=j}\overline{\mu}_i(x)\sup_{x'_j \in B_p(x,\gamma)} \Phi_{\mathrm{cls}}^{\rho,u}\big(h(x'_j),j\big)
    \end{aligned}
\end{equation}
Let's define the probability distribution $p_j(x) = \sum_{i\not=j}\overline{\mu}_i(x)/S(x)$ with $S(x) =\sum_{j=1}^{K+J}\sum_{k\not=j}\overline{\mu}_k(x)$. It follows the calibration gap and by Lemma \ref{lemma:qconsistency}:
\begin{equation}
    \begin{aligned}
        \Delta\mathcal{C}_{\widetilde{\Phi}_{\mathrm{def}}^{c,u}}(h,x) & = S(x) \Bigg( \sum_{j=1}^{K+J} p_j(x)\sup_{x'_j \in B_p(x,\gamma)} \Phi_{\mathrm{cls}}^{\rho,u}\big(h(x'_j),j\big) - \inf_{h \in \mc{H}} \sum_{j=1}^{K+J}p_j(x)\sup_{x'_j \in B_p(x,\gamma)} \Phi_{\mathrm{cls}}^{\rho,u}\big(h(x'_j),j\big)\Bigg) \\
        & \geq S(x) [\Psi^u(1)]^{-1} \Bigg( \sum_{j=1}^{K+J} p_j(x)\sup_{x'_j \in B_p(x,\gamma)} \mathbf{1}\{\hat{h}(x'_j)\not=j\}\big) - \inf_{h \in \mc{H}} \sum_{j=1}^{K+J}p_j(x)\sup_{x'_j \in B_p(x,\gamma)} \mathbf{1}\{\hat{h}(x'_j)\not=j\}\big)\Bigg)
    \end{aligned}
\end{equation}

\paragraph{Relationship between Calibration gaps.}
We fix \(h\in\mathcal H\) and analyze its excess risks at a given \(x\). Define
\[
\mathcal E_p(h,x)=\sum_{j=1}^{K+J}p_j(x)\,\sup_{x'_j\in B_p(x,\gamma)}\mathbf 1\{\hat h(x'_j)\neq j\}.
\]
Write for conciseness
\[
\mu_\ast(x)=\min_{j\in\mathcal A}\overline\mu_j(x),
\]
and use
\[
M(x)=\sum_{i=1}^{K+J}\overline\mu_i(x),
\qquad
S(x)=\sum_{j=1}^{K+J}\sum_{k\neq j}\overline\mu_k(x)=(K{+}J{-}1)\,M(x),
\]
so that \(p_j(x)=\dfrac{M(x)-\overline\mu_j(x)}{S(x)}\).
For a fixed \(h\), \(\sup_{x'_j}\mathbf 1\{\hat h(x'_j)\neq j\}=0\) iff \(j\in \mc{H}_\gamma(h,x)\) and
\(\hat h(x')=j\) for all \(x'\in B_p(x,\gamma)\), and equals \(1\) otherwise. Hence
\[
\mathcal E_p(h,x)
=1-\sum_{j\in\mathcal A}p_j(x)\,\mathbf 1\big\{j\in \mc{H}_\gamma(h,x)\ \text{and}\ \hat h \text{ is constant }j\text{ on }B_p(x,\gamma)\big\}.
\]
Allowing a predictor that is constant on the ball with the most likely \(p\)-class yields the Bayes value
\[
\mathcal E_p^\star(x)=1-\max_{j\in\mathcal A}p_j(x)=1-\frac{M(x)-\mu_\ast(x)}{S(x)}.
\]
Therefore,
\begin{equation}
\label{eq:robust-excess-identity-star}
\mathcal E_p(h,x)-\mathcal E_p^\star(x)
=\max_{j}p_j(x)\;-\;\sum_{j\in\mathcal A}p_j(x)\,\mathbf 1\big\{j\in \mc{H}_\gamma(h,x)\ \text{and}\ \hat h \text{ is constant }j\text{ on }B_p(x,\gamma)\big\}.
\end{equation}

We therefore have two different cases to inspect. 

\emph{Case (i): \(h\) is constant on \(B_p(x,\gamma)\) with label \(j_0\).}

Then the RHS of \eqref{eq:robust-excess-identity-star} equals \(\max_j p_j(x)-p_{j_0}(x)\).
Using \(p_j(x)=\frac{M(x)-\overline\mu_j(x)}{S(x)}\) and that \(\max_j p_j\) occurs at any minimizer of \(\overline\mu_j\),
\[
\mathcal E_p(h,x)-\mathcal E_p^\star(x)=\frac{\overline\mu_{j_0}(x)-\mu_\ast(x)}{S(x)}.
\]
Since here \(\mc{H}_\gamma(h,x)=\{j_0\}\), the true excess is
\(\Delta\mathcal C_{\widetilde{\ell}_{\mathrm{def}}^c}(h,x)=\overline\mu_{j_0}(x)-\mu_\ast(x)\), hence
\begin{equation}
\label{eq:key-bound-case1-star}
\mathcal E_p(h,x)-\mathcal E_p^\star(x)=\frac{1}{S(x)}\,\Delta\mathcal C_{\widetilde{\ell}_{\mathrm{def}}^c}(h,x).
\end{equation}

\emph{Case (ii): \(h\) is not constant on \(B_p(x,\gamma)\).}

Then the second term in \eqref{eq:robust-excess-identity-star} vanishes and
\[
\mathcal E_p(h,x)-\mathcal E_p^\star(x)=\max_j p_j(x)=\frac{M(x)-\mu_\ast(x)}{S(x)}.
\]
Since \(\sum_{j\in \mc{H}_\gamma(h,x)}\overline\mu_j(x)\le M(x)\), we obtain
\begin{equation}
\label{eq:key-bound-case2-star}
\mathcal E_p(h,x)-\mathcal E_p^\star(x)\ \ge\ \frac{\sum_{j\in \mc{H}_\gamma(h,x)}\overline\mu_j(x)-\mu_\ast(x)}{S(x)}
=\frac{1}{S(x)}\,\Delta\mathcal C_{\widetilde{\ell}_{\mathrm{def}}^c}(h,x).
\end{equation}

Combining \eqref{eq:key-bound-case1-star}–\eqref{eq:key-bound-case2-star} yields, for all \(h\in\mathcal H\),
\begin{equation}
\label{eq:robust01-vs-true-star}
\mathcal E_p(h,x)-\mathcal E_p^\star(x)\ \ge\ \frac{1}{S(x)}\,\Delta\mathcal C_{\widetilde{\ell}_{\mathrm{def}}^c}(h,x).
\end{equation}

\paragraph{Lower-Bounding the Surrogate Calibration Gap.} Using (\ref{eq:robust01-vs-true-star}), we obtain:
\begin{equation}
    \begin{aligned}
        \Delta\mathcal{C}_{\widetilde{\Phi}_{\mathrm{def}}^{c,u}}(h,x) & = S(x) \Bigg( \sum_{j=1}^{K+J} p_j(x)\sup_{x'_j \in B_p(x,\gamma)} \Phi_{\mathrm{cls}}^{\rho,u}\big(h(x'_j),j\big) - \inf_{h \in \mc{H}} \sum_{j=1}^{K+J}p_j(x)\sup_{x'_j \in B_p(x,\gamma)} \Phi_{\mathrm{cls}}^{\rho,u}\big(h(x'_j),j\big)\Bigg) \\
        & \geq S(x) [\Psi^u(1)]^{-1} \Bigg( \sum_{j=1}^{K+J} p_j(x)\sup_{x'_j \in B_p(x,\gamma)} \mathbf{1}\{\hat{h}(x'_j)\not=j\}\big) - \inf_{h \in \mc{H}} \sum_{j=1}^{K+J}p_j(x)\sup_{x'_j \in B_p(x,\gamma)} \mathbf{1}\{\hat{h}(x'_j)\not=j\}\big)\Bigg) \\
        & \geq S(x) [\Psi^u(1)]^{-1} \Bigg(\frac{1}{S(x)}\,\Delta\mathcal C_{\widetilde{\ell}_{\mathrm{def}}^c}(h,x)\Bigg) \\
        & = [\Psi^u(1)]^{-1} \Bigg(\Delta\mathcal C_{\widetilde{\ell}_{\mathrm{def}}^c}(h,x)\Bigg)
    \end{aligned}
\end{equation}

Applying the expectation $\mb{E}_X[\Delta\mc{C}_{\ell}(h,X)] = \mathcal{E}_{\ell}(h) - \mathcal{E}_{\ell}^B(\mc{H}) + \mathcal{U}_{\ell}(\mc{H})$, leads to the desired inequality:
\begin{equation*}
    \begin{aligned}
        & \mathcal{E}_{\widetilde{\ell}^c_{\text{def}}}(h) - \mathcal{E}_{\widetilde{\ell}^c_{\text{def}}}^B(\mc{H}) + \mathcal{U}_{\widetilde{\ell}^c_{\text{def}}}(\mc{H})  \leq \Psi^u(1) \Big( \mathcal{E}_{\widetilde{\Phi}^{c,u}_{\text{def}}}(h) - \mathcal{E}_{\widetilde{\Phi}^{c,u}_{\text{def}}}^\ast(\mc{H}) + \mathcal{U}_{\widetilde{\Phi}^{c,u}_{\text{def}}}(\mc{H}) \Big),
    \end{aligned}
\end{equation*}
with $\Psi^u(1) =
\begin{cases}
\log(2), & u=1, \\[0.3em]
\tfrac{1}{1-u}\big(2^{1-u}-1\big), & u \neq 1.
\end{cases}$

\end{proof}

\subsection{Proof of Lemma \ref{lem:tlcrobustness-reg}} \label{proof:lem:tlcrobustness-reg}

\tlcrobustnessreg*

\begin{proof}
We begin by recalling the standard true deferral loss for regression from \citet{mao2024regressionmultiexpertdeferral}: 
\begin{equation}
    \ell_{\mathrm{def}}^r\big(\hat{r}(x), f(x), t, \mathbf{m}\big) 
    = \sum_{j=1}^{J+1} c_j^r\big(f(x), m_{j-1}, t\big)\,
      \mathbf{1}\{\hat{r}(x) = j\}.
\end{equation}
To capture the worst-case scenario, we account for adversarial perturbations of the input while noting that 
the cost $c_1^r(f(x),t)$ depends explicitly on $x$ and that the predictor $f \in \mathcal{F}$ is trainable: 
\begin{equation}
    \begin{aligned}
        \ell_{\mathrm{def}}^r\big(\hat{r}(x), f(x), t, \mathbf{m}\big) 
        &\leq \sup_{x' \in B_p(x,\gamma)} 
            \sum_{j=1}^{J+1} c_j^r\big(f,x, m_{j-1}, t\big)\,
            \mathbf{1}\{\hat{r}(x') = j\} \\[0.5em]
        &\leq \sum_{j=1}^{J+1} \tilde{c}_j^r\big(f,x, m_{j-1}, t\big)\,
            \sup_{x'_j \in B_p(x,\gamma)} \mathbf{1}\{\hat{r}(x'_j) = j\} \\[0.5em]
        &= \tilde{\ell}_{\mathrm{def}}^r\big(r, f, x, t, \mathbf{m}\big).
    \end{aligned}
\end{equation}

\end{proof}

\subsection{Proof of Theorem \ref{theo:r_consistency}} \label{proof:theo:r_consistency}

\rconsistency*

\begin{proof}
The outcome-wise adversarial deferral loss for regression is defined as
\begin{equation}
 \begin{aligned}
    \tilde{\ell}^r_{\mathrm{def}}(f,r,x,t,\mathbf{m})
    = \sum_{j=1}^{J+1} \tilde{c}_j^r(f,x,m_{j-1},t)\,
      \sup_{x_j' \in B_p(x,\gamma)} \mathbf{1}\{\hat{r}(x_j') = j\},
 \end{aligned}
\end{equation}

\end{proof}

\subsection{Complexity}\label{complexity}
\begin{proposition}[Epoch cost of RERM-C in the $h$-score setting]
\label{prop:epoch-cost-h}
Process $n$ training examples in mini-batches of size $B$.
Let $\mathcal A$ be action space
(\,e.g.\ $\mathcal A=\{1,\dots,K{+}J\}$\,) and let $T\in\mathbb N$ be the
number of projected-gradient steps used in each inner maximization
(\textsc{PGD}($T$)).  Write $C_{\mathrm{fwd}}$ and $C_{\mathrm{bwd}}$
for the costs of one forward and one backward pass through the score
network $h$.
Then one epoch of RERM-C minimization incurs
\begin{equation}
\label{eq:epoch-cost}
n\,\bigl(1+|\mathcal A|\,T\bigr)\,\bigl(C_{\mathrm{fwd}}+C_{\mathrm{bwd}}\bigr)
\end{equation}
network traversals, while the peak memory equals that of a single
forward–backward pass plus the storage of one adversarial copy of each
input currently being optimized.
\end{proposition}

\begin{proof}
Consider one mini-batch. RERM-C performs:
\begin{enumerate}\itemsep0pt
\item[(i)] one \emph{clean} forward pass of $h$ to compute scores $h(x)$ and the loss terms;
\item[(ii)] for each $j\in\mathcal A$ and each of the $T$ \textsc{PGD} steps that update the
            adversarial proxy $x'_j\in B_p(x,\gamma)$, one forward \emph{and} one backward
            pass of $h$ (to obtain gradients w.r.t.\ the input);
\item[(iii)] one backward pass to update $\theta$ (the parameter gradient of the batch loss).
\end{enumerate}
Thus, per mini-batch, the total number of network traversals equals
\[
\underbrace{1}_{\text{clean forward}}
+\underbrace{|\mathcal A|\,T}_{\text{\textsc{PGD} forwards}}
+\underbrace{|\mathcal A|\,T}_{\text{\textsc{PGD} backwards}}
+\underbrace{1}_{\text{parameter backward}}
\;=\;2\bigl(1+|\mathcal A|\,T\bigr).
\]
Multiplying by the number of mini-batches $\lceil n/B\rceil$ gives the total
cost
\[
\lceil n/B\rceil \cdot 2\bigl(1+|\mathcal A|\,T\bigr)
\;\approx\; \frac{n}{B}\cdot 2\bigl(1+|\mathcal A|\,T\bigr),
\]
which, when expressed in units of \emph{per-example} forward/backward costs,
yields \eqref{eq:epoch-cost}:
each example induces $\bigl(1+|\mathcal A|\,T\bigr)$ forwards and the
same number of backwards, for a total of
$n\bigl(1+|\mathcal A|\,T\bigr)\bigl(C_{\mathrm{fwd}}+C_{\mathrm{bwd}}\bigr)$.

\end{proof}

\subsection{Experiments}

\subsubsection{Resources} \label{ressources}
All experiments were conducted on an internal cluster using an NVIDIA A100 GPU with 80 GB of VRAM.

\subsubsection{CIFAR10} \label{cifar10_exp}

\begin{table}[H]
\centering
\begin{tabular}{c c c c }
\toprule
\textbf{Expert} & 1 & 2 & 3  \\
\midrule
\textbf{Accuracy} & 37.55 & 35.92 & 38.54  \\
\bottomrule
\end{tabular}
\caption{CIFAR10: Accuracy of Experts}
\end{table}
Analysis in the main paper. 

\subsubsection{DermaMNIST}\label{dermnist_app}

\begin{table}[h!]
\centering
\begin{tabular}{c c c c}
\toprule
\textbf{Expert} &  1 & 2 & 3 \\
\midrule
\textbf{Accuracy} & 28.48 & 30.52 & 71.72\\
\bottomrule
\end{tabular}
\caption{DermaMNIST: Accuracy of Experts}
\end{table}

\phantomsection
\paragraph*{Setting.} 
DermaMNIST is a subset of the MedMNIST dataset consisting of biomedical images for 7-class classification. 
The augmented classifier is implemented using ResNet-18 and trained for 100 epochs. 
As experts, we construct three specialized classifiers, each responsible for a randomly assigned subset of three classes (with overlap), predicting correctly with probability $p=0.85$ on their assigned classes and uniformly at random otherwise; their accuracies are reported above. 
The consultation costs are set as $\beta_{j \leq K}=0$ for predictions, and $\beta_{K+1}=0.05$, $\beta_{K+2}=0.075$, and $\beta_{K+3}=0.125$ for the experts. 
Both the baseline method and our approach use a learning rate of $0.005$. 
For the PGD attack, we set $\epsilon=0.03137$, while our approach additionally uses the hyperparameters $\rho=1.75$ and $\nu=0.001$.

\phantomsection
\paragraph*{Results.}
\begin{table}[ht]
  \centering
    \begin{tabular}{@{}lcccccc@{}}
      \toprule
       & C.Acc & U.Acc & T.Acc & Def.Loss  \\
      \midrule
      \citet{mao2024principledapproacheslearningdefer}  & $83.39$ & $30.82$ & $27.08$ & $69.60$  \\
      Ours &  $81.80$ & $71.12$ & $80.65$ & $31.66$  \\
      \bottomrule
    \end{tabular}%
  \caption{Performance under clean and adversarial inputs, compared against the approach of \citet{mao2024principledapproacheslearningdefer}.}
  \label{tab:dermamnist_app}
\end{table}

On the DermaMNIST dataset, Table~\ref{tab:dermamnist_app} shows that our approach remains close to the baseline under the clean setting while substantially improving targeted and untargeted attacked accuracy, together with a lower empirical adversarial deferral loss.

\subsubsection{Community and Crime}

\begin{table}[H]
\centering
\begin{tabular}{c c c c }
\toprule
\textbf{Expert} & 1 & 2 & 3   \\
\midrule
\textbf{RMSE} & 0.5442 & 1.1373 & 1.5613  \\
\bottomrule
\end{tabular}
\caption{community and crime: Accuracy of Experts (RMSE)}
\end{table}

Analysis in the main paper. 

\subsubsection{Insurance Company Benchmark (COIL 2000)}
\begin{table}[H]
\centering
\begin{tabular}{c c c c c}
\toprule
\textbf{Expert} & 1 & 2 &3 & 4  \\
\midrule
\textbf{RMSE} & 0.0744 & 0.0747 & 0.0741 & 0.831\\
\bottomrule
\end{tabular}
\caption{coil2000: Accuracy of Experts (RMSE)}
\end{table}
\phantomsection
\paragraph*{Setting.}

The rejector is implemented using an MLP and trained with the AdamW optimizer \citep{kingma2017adammethodstochasticoptimization} for 25 epochs, while the main predictor is a linear layer. As experts, we employ four regression MLPs, each focusing on different customer segments (demographics, product ownership, high-value customers) and generating predictions using rules and noise; their accuracies are reported above. The consultation costs are set as follows: $\beta_{1}=0$ for the main predictor, and $\beta_2=0.035$, $\beta_3=0.04$, $\beta_4=0.045$ and $\beta_5=0.05$ for the experts. The baseline method uses a learning rate of $0.005$, while our approach employs a learning rate of $0.01$. For the PGD attack, we set $\epsilon=2$, and in our method we additionally use the hyperparameters $\rho=2.75$ and $\nu=0.01$.

\phantomsection
\paragraph*{Results.}

\begin{table}[ht]
  \centering
    \begin{tabular}{@{}lcccccc@{}}
      \toprule
      & C.Acc & U.Acc & T.Acc & Def.Loss \\
      \midrule
      \citet{mao2024regressionmultiexpertdeferral} &  $7.02$ & $11.61$ & $8.31$ & $11.98$  \\
      Ours &  $7.39$ & $7.41$ & $7.40$ & $7.81$  \\
      \bottomrule
    \end{tabular}%
  \caption{Performance under clean and adversarial inputs, compared against the approach of \citet{mao2024regressionmultiexpertdeferral}.}
  \label{tab:coil_appendix}
\end{table}

On the COIL-2000 dataset, our approach remains close to the baseline under the clean setting while improving both attacked RMSE and adversarial deferral loss, suggesting that the method transfers reasonably well across the reported regression benchmarks.

\end{document}